# Staircase Recognition and Location Based on Polarization Vision

Weifeng Kong, Zhiying Tan

***Abstract*— Staircase perception is critical for humanoid robots and mobility-impaired individuals, yet existing methods have low accuracy, lighting sensitivity, and texture dependency. To address this, we propose a novel polarization-visual fusion framework that achieves robust staircase detection and high-precision the three-dimensional (3D) reconstruction, establishing a paradigm of Staircase recognition — Heterogeneous sensor calibration (monocular and TOF camera) — Polarization 3D reconstruction. First, the staircase recognition algorithm based on YOLOv11 integrated with polarization-intensity contrast enhancement algorithm and point cloud segmentation is improved, reaching recognition accuracy of 98.7%±0.10% by suppressing reflections and correcting by redundant information of point cloud. Then, an improved gray wolf optimizer with Levy flight and dynamic weights enables accurate heterogeneous sensor calibration (0.33±0.04 mm error) between heterogeneous-resolution cameras is employed. Finally, a method of fusing polarized binocular and TOF depth information to realize the 3D reconstruction of the staircase is proposed. Considering the ambiguity in polarization reconstruction and the data holes in binocular reconstruction, binocular vision is used to correct polarization azimuth ambiguity, TOF is used to fill data holes from stereo matching. Experiments show our method achieves <0.2% reconstruction error at 0.5m – significantly outperforming binocular (surface distortion) and polarization-based (normal vector ambiguity) approaches. This technology provides accurate terrain adaptation for robotic foothold planning.**

## I. INTRODUCTION

As a general scene, the staircase interferes with the traversal of humanoid robots, legged robots, lower limb disabilities, or visually impaired individuals due to its special physical structure. Accurate staircase recognition technology is a prerequisite for navigation and control, and staircase recognition technology has attracted the attention of many scholars [1],[2],[3]. Staircase recognition is of great significance for the mode switching and foothold position calculation of robots, which can improve the overall performance of robots in staircase scenes. As a common terrain, stairs are very difficult for humanoid robots and people with lower limb disabilities or visual impairments. Therefore, it is of great significance to design a staircase scene perception algorithm.

At present, the staircase recognition is mainly applied in the fields of rehabilitation medicine and humanoid robots [4]. Because accurate gait control is one of the prerequisites for normal gait, while gait disorders can lead to cessation of movement, falls, and increased mortality. The accurate recognition of human gait requires the joint action of sensors and signal processing algorithms [5], and the signals collected by sensors are processed through the upper computer [6]. This technology can provide auxiliary guidance to target populations, such as visual navigation systems for blind individuals and gait monitoring for Parkinson's patients. The sensors used for staircase recognition are mainly divided into wearable sensors and photoelectric sensors. Wearable sensors include inertial measurement units, electromyographic sensors, etc.

Wearable sensors are mainly aimed at recognizing the intention of movement in staircase scenes. The signals collected by the sensors are processed through classification algorithms or neural networks to achieve the classification of motion intentions, thereby achieving the goal of indirect scene recognition [7]. Recent studies have demonstrated significant advancements in motion recognition through the integration of neural networks with inertial and physiological signals. Hernandez applicated neural networks in this domain by developing a hybrid model comprising a convolutional neural network (CNN), a predictive information gain module, and an adaptive information fusion mechanism for IMU data analysis, achieving an exceptional recognition accuracy of 99.9% [8]. Building upon this, Coelho's team introduced a real-time gait detection algorithm that synergistically combines CNN and LSTM architectures. Their system effectively captures gait patterns across diverse terrains, including level walking, staircase ascent, and descent, enabling prompt adaptation to environmental transitions [9]. For practical implementation, Chen's work demonstrated the feasibility of deploying machine learning algorithms on mobile platforms, utilizing an integrated extreme learning machine approach for human activity via smartphone IMU sensors [10]. In addressing the challenges of electromyography signal processing, researchers have developed sophisticated noise-robust classification frameworks. Xi's approach employed wavelet transformation of preprocessed surface electromyography signals, with the derived coherence coefficients serving as input to an SVM classifier, achieving reliable identification of stair-related movements [11]. To further enhance computational efficiency, subsequent studies incorporated dimensionality reduction techniques for wavelet-transformed EMG features. This optimization enabled the implementation of a novel lower limb motion classification system combining scale-unscented Kalman filtering with neural networks, where network parameters were refined through scale-corrected unscented transformation [12]. Although wearable sensors have applications in recognizing motion intentions, they possess inherent limitations. Inertial measurement unit data may be distorted due to body movements (such as limb swinging), leading to inaccurate classification of motion intentions. Surface electromyography signals inherently contain noise, necessitating complex denoising techniques (such as wavelet transform) to ensure their usability. Furthermore, they rely on interpreting motion intentions to deduce scene characteristics, which can result in delays and potential mismatches between expected and actual scenes.

In contrast, photoelectric sensors directly capture visual or depth information of the environment (such as images, point clouds), leveraging their unique advantages for direct scene perception and enabling clear recognition of staircase without relying on intermediate motion signals. Photoelectric sensors include monocular camera, depth camera, radar and so on, which are mainly used for scene recognition before entering the staircases. Staircase recognition is very important in robot motion navigation and control, so it is necessary to recognize the staircase, but the existing algorithms have the problems of low recognition accuracy, slow running speed and unable to achieve scene versatility [13]. Recent studies have successfully applied the YOLO algorithm, a widely-used object detection method, to staircase recognition by integrating it with Hough transform techniques. This approach first identifies regions of interest through YOLO detection, then determines movement directions using Hough line detection within these regions [14-16]. Yang further enhanced this methodology by incorporating it into a wearable navigation system with deep segmentation capabilities, validating its effectiveness through real-world tests with visually impaired users [17]. Nevertheless, monocular imaging systems face limitations in providing precise dimensional data for robotic navigation in staircase environments. To address this challenge, Westfechtel developed a plane segmentation-based approach that identifies planar trapezoids and their ascending components, creating a dynamic graphical model for comprehensive staircase and track system recognition [18]. Meanwhile, Jia proposed an alternative solution using depth data in inverse depth coordinates, employing an iterative RANSAC-based segmented fitting method that incorporates inclination analysis [19]. While 3D recognition methods can capture detailed structural information, they demand significant computational resources. The present study strikes an optimal balance between accuracy and efficiency by implementing a "2D image recognition first, point cloud verification second" strategy.

To enhance the robot's motion capabilities, 3D reconstruction of the staircase based on the recognition results is indispensable. Nowadays, various sensors are widely used in the 3D reconstruction, but these sensors have the characteristics of poor imaging details, low imaging accuracy and efficiency. For example, line structured light cannot achieve single-frame shape recovery because its projection pattern is a straight line, resulting in low efficiency [20]. The projected structured light is easily affected by reflection, illumination and other reasons, and the encoded information is lost in some areas. Depth camera and radar are limited by factors such as time resolution, environment, and multipath effects [21]. The multipath effect of depth sensor data is shown in Fig.1, which causes discontinuous depth regions in the depth map, resulting in errors in depth measurement. To solve this problem, a multi-sensor fusion method is introduced to fuse polarization and TOF for 3D reconstruction of staircase.

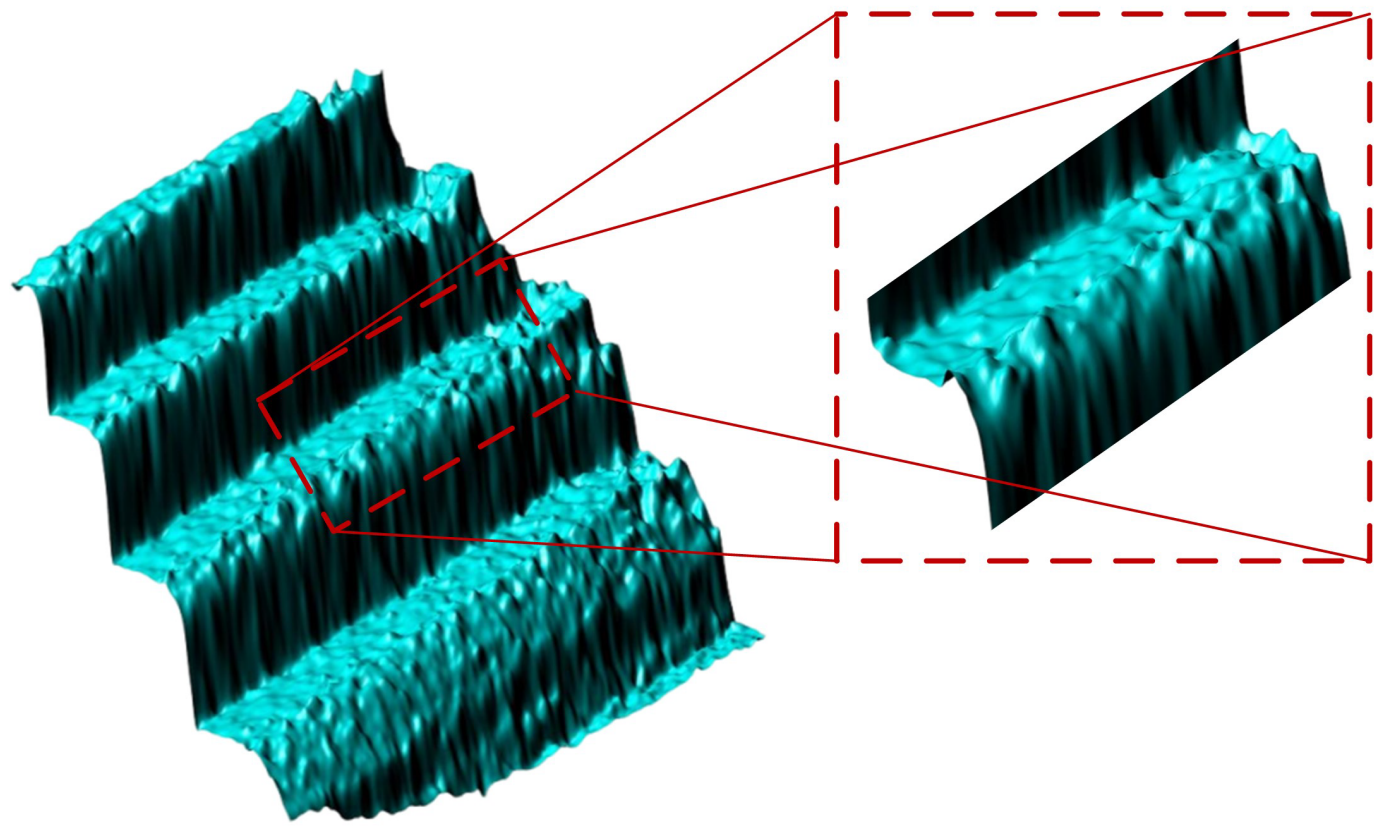

**Fig.1 Schematic diagram of multipath effect in depth sensor data**

Polarization imaging can obtain more multi-dimensional information besides the ordinary intensity information [22]. Reflection of unpolarized light by the surface of an object changes the polarization of the light, and the polarization of the light reflected from the object conveys information about the material properties and shape of the object. 3D reconstruction can be carried out by detecting the degree of polarization (DoP) and polarization angle of the object surface. Polarization 3D reconstruction technology has the advantages of simple imaging equipment, easy to implement algorithm, rich reconstruction details and so on, so polarization 3D reconstruction technology has important research significance [23]. Nevertheless, there are a series of problems, such as the ambiguity of the azimuth angle and the inability to obtain the real 3D information of the object [24],[25]. Previous approaches have attempted to address these limitations through various methods. Liu employed shape-from-shading techniques to correct azimuth ambiguity by estimating surface normal vectors from brightness variations, assuming that the light source is a uniform parallel light or a point light source with a known position [26]. However, in real-world scene, illumination is often uneven, leading to deviations in the distribution of brightness and darkness on the object surface from the theoretical model [27]. In the research process of Cai, he developed a feature-based approach that resolves azimuth ambiguity using prior knowledge of target image characteristics. Specifically, through statistical analysis of the probability distribution of true azimuth angles, it was found that their quadrant distribution is closely related to the position of facial feature points. Therefore, by detecting facial features, the polarization normal angle of each pixel can be accurately assigned to the corresponding quadrant, thereby determining the precise unique normal vector and achieving accurate 3D face reconstruction. Although, this method can overcome azimuth ambiguity without extra depth devices, this method relies on the prior correlation between facial feature points and the distribution of azimuth quadrants. Therefore, this method cannot be directly applied to staircase reconstruction due to the lack of distinct feature points in low-texture planar regions, resulting in a loss of basis for azimuth quadrant allocation [28]. Other researchers have explored hybrid sensing techniques. Tan combined polarization imaging with structured light to reconstruct 3D coordinates. The 3D coordinates of the laser fringe and the relative depth value of the polarization three-dimensional reconstruction are combined to calculate the real point cloud of the target object, which needed to capture multiple images at static condition and is not suitable for our mobile robot application scene [29]. Cui proposed a multi-view geometric constraint approach using polarization information, but rapid perspective changes during robot motion make stable data acquisition challenging [30]. Recently, the development of deep learning has made great contributions in the area of ambiguity correction. Han integrated photometric and geometric cues from the polarimetric image formation model and multiview azimuth consistency with implicit neural representation. However, this method requires paired polarization-depth datasets through high-precision scanners, that are difficult to obtain in outdoor mobile applications. [31].

To solve these problems, here we integrated binocular polarization vision with TOF, realizing the fusion of active and passive imaging. Compared with depth camera, radar and monocular camera, binocular vision can simultaneously obtain the appearance image and three-dimensional information of an object. It can not only utilize the geometric features of the object, but also analyze and correct the ambiguity of the polarization gradient field by combining its visual features such as texture and color, providing more comprehensive and rich information. However, data holes are inevitable in the stereo matching process [32]. Although the TOF method is affected by multipath effects, it can obtain full pixel reconstruction [33]. The fusion strategy significantly advances polarization-based reconstruction by offering three key advantages over previously discussed methods: robust performance on low-texture surfaces like staircase treads independent of specific surface features (overcoming limitations of feature-dependent methods [28]), single-frame imaging capability enabling operation in dynamic mobile robot scene (unlike multi-frame static approaches [29]),and no requirement for paired ground-truth depth data eliminating the need for difficult-to-acquire high-precision scanner datasets [31]. Therefore, combining these methods, this paper uses binocular to correct polarization ambiguity and TOF to correct binocular hole regions. Their complementary advantages not only correct ambiguity, but also effectively avoid data hole regions. However, the fusion of multiple sensors will face the problem of sensor calibration, arising from the significant resolution disparity between the sensors, constitutes a key focus and contribution of this paper. To the best of our knowledge, this represents the first international effort towards 3D reconstruction utilizing the fusion of binocular polarization vision and TOF sensing.

To sum up, in order to establish a paradigm of Staircase recognition — Heterogeneous sensor calibration — Polarization 3D reconstruction, improvements of each module are as follows:

(1) Staircase Recognition Module: A staircase recognition method based on polarization information fusion and 2D-3D fusion is proposed based on Multilayer Perceptron (MLP) network, which significantly enhances image contrast through feature level fusion of DoP image and intensity image. Then, point cloud normal vector segmentation assisted recognition further improves the detection accuracy and robustness of staircase.

(2) Heterogeneous Sensor Calibration Module: Since polarization camera and TOF camera have different resolutions, so the traditional binocular calibration algorithm cannot be applied in this condition. We proposed a joint calibration algorithm combined with the improved joint grey wolf optimization algorithm with Levy flight and dynamic weights for ICP algorithm.

(3) Polarization 3D Reconstruction Module: Polarization gradient field ambiguity is corrected by fusion binocular and TOF. Firstly, a three-dimensional shape similar to the surface of the target object is reconstructed by a binocular vision method, and a relatively accurate gradient field is obtained, which is used as a reference to correct the ambiguity problem in the polarization gradient field. Due to the existence of data holes in the binocular stereo matching process, the polarization gradient field corresponding to this part of the region has not been effectively corrected, and the TOF depth information is introduced to correct it.

## II. Staircase detection based on polarization information fusion and 2D-3D fusion

In the motion of robot staircase scenes, real-time reconstruction of the environment requires a large amount of computing resources. In order to save computing power, the staircase can be recognized to determine the ROI area of the staircase in the field of view, and then trigger scene reconstruction. Reconstructing scenes within local ROI regions not only avoids the consumption of computing resources, but also improves real-time performance.

In object detection tasks, the reflection on the surface of the target object is one of the key factors affecting detection accuracy. Reflection can cause abnormal enhancement of pixel intensity in the target region, thereby disrupting the texture and edge information of the image and reducing the contrast between the target and the background. This makes it difficult for object detection models to accurately distinguish between target and non-target regions, thereby reducing the accuracy and robustness of detection.

The DoP image is an image generated by measuring the polarization state of light. It can effectively capture the polarization information of the target surface, highlight the texture and edge features of objects, and simultaneously suppress the reflective regions, thereby enhancing the contrast and detectability of the image [34]. The DoP image can be calculated from images $I_0$, $I_{45}$, $I_{90}$, and $I_{135}$ using the Stokes method. When the polarization angles are 0 °, 45 °, 90 °, and 135 °, respectively. DoP is calculated as follows:

$$\rho = \frac{\sqrt{{s_1}^2 + {s_2}^2}}{s_0} \tag{1}$$

where, $s_0 = (I_0 + I_{45} + I_{90} + I_{135}) / 4$, $s_1 = I_0 - I_{90}$, $s_2 = I_{45} - I_{135}$.

The intensity image intuitively displays the distribution of light intensity and can clearly present the contours, shapes, and general surface features of objects. To address the issue of reflection, a fusion strategy for the DoP image and the intensity image based on a MLP is proposed. By integrating these two maps, the detailed information provided by the DoP image, such as microstructure, can be incorporated into the overall contour of the intensity image. This allows us to grasp the overall shape of the object while also observing its fine details, thereby obtaining richer and more comprehensive image information.

The structure of the DoP and intensity image fusion network is shown in Fig.2, which is located at the input of YOLOv11. Convolution operations are first performed on the data of the two channels, namely the DoP and the intensity image, to extract local features and enhance the feature representation. Subsequently, the data from the two channels after convolution processing are concatenated at the element level to form an integrated feature vector, effectively fusing the information of polarization and intensity, and providing a more comprehensive feature representation for subsequent analysis. This feature vector is then input into a Multilayer Perceptron (MLP) network for training and learning. Leveraging its powerful feature learning capabilities, the MLP network can automatically identify features and patterns in the data and output two weights, $DoP_{weight}$ and $I_{weight}$, corresponding to the importance of the DoP image and the intensity image in the fusion process, respectively. Finally, these weights are then multiplied element-wise with their respective images and summed to achieve image fusion. The final fused image can be represented as:

$$I_{final} = DoP_{weight} * P + I_{weight} * I \tag{2}$$

where, $P$ and $I$ represent DoP image and intensity image, respectively.

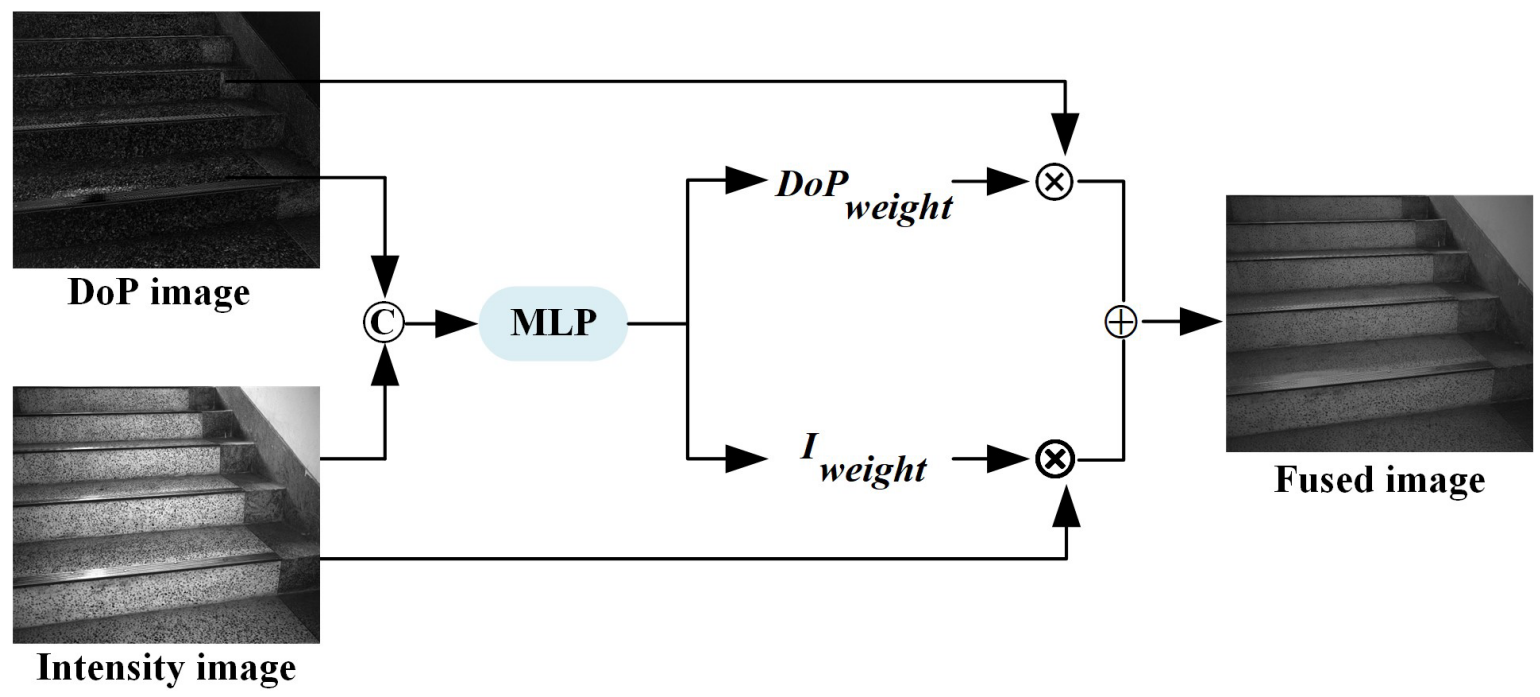

**Fig.2 Polarization and intensity image fusion based on MLP network**

This weighted fusion method fully considers the characteristics of the two images: the DoP image can effectively capture the polarization information of the target surface, while the intensity map retains the brightness information of the original image. Through the non-linear mapping capability of the MLP network, the fusion of these two types of image information at the feature level significantly enhances the contrast of the fused image.

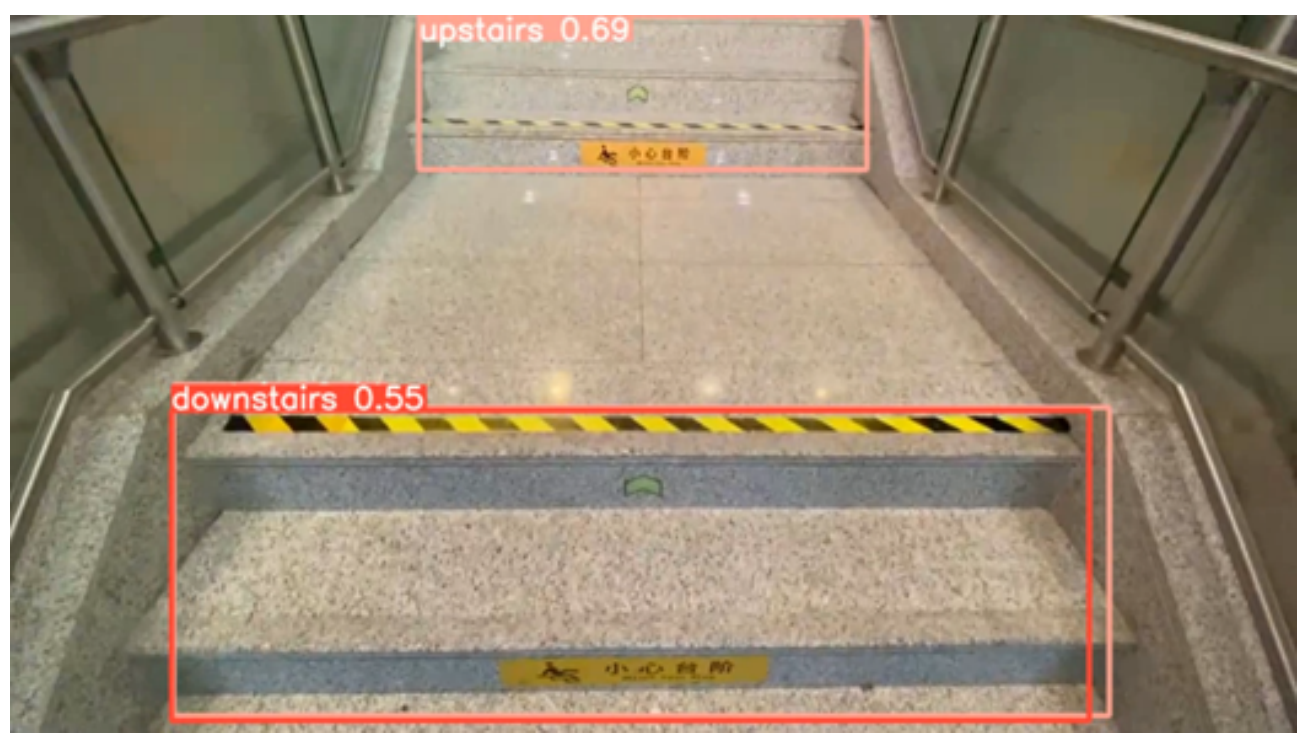

**Fig.3 Recognition error example**

The differences and advantages of YOLOv11 compared to YOLOv5 and YOLOv7 are mainly reflected in the following aspects: Firstly, the backbone network adopts an improved CSP structure, which reduces parameters by 15%-20% through dynamic channel pruning and enhances the ability to capture multi-scale features, making it more sensitive to details such as staircase edges. In contrast, the structure of YOLOv5 is relatively fixed, and YOLOv7 is slightly inferior in retaining small target features. Secondly, the neck part improves the PAN-FPN by introducing adaptive weighted fusion, which solves the problem of feature imbalance in YOLOv5 and redundant interference in YOLOv7, enhances robustness in complex environments, and is conducive to integrating the information of DoP images and intensity images in staircase scenes. Thirdly, it adopts a pure Anchor-Free detection head and CIoUv2 loss, reducing dependence on anchor boxes and improving the positioning accuracy of step edges, which is superior to the Anchor-Based of YOLOv5 and the hybrid detection head of YOLOv7. Although the optimized YOLOv11 model has shown satisfactory performance in practical applications, recognition errors are still inevitable. As shown in Fig.3, the model encountered a misjudgment when distinguishing between upstairs and downstairs, mistakenly identifying the scene upstairs as downstairs. In order to improve the robustness and accuracy of the algorithm, this paper proposes a method of using depth camera to assist recognition. Depth camera can capture the 3D geometric information of objects in the scene, thus providing more abundant and accurate data for going up and down stairs. By combining the depth camera data, we can verify or correct the recognition results of YOLOv11.The staircase point cloud is divided into vertical and horizontal planes by using the difference of normal differences, and further recognition analysis is carried out according to the probability distribution of the two planes.

According to the characteristics of most staircases, they are similar to those in Fig.4. The horizontal planes are labeled "1", "2", "3", and the vertical planes are labeled "a", "b", "c". Theoretically, there is a normal line perpendicular to the surface as shown in Fig.4. This structure motivates us to segment the staircase surface by KD-Tree Euclidean clustering according to the distribution of normal vectors $\tilde{n}_1$ and $\tilde{n}_2$ on the vertical and horizontal planes of the staircase surface, which is inspired by the difference of the direction of the normal vector of the staircase surface $\theta_{\text{diff}} = \cos^{-1}(\vec{n}_i \cdot \vec{n}_j)$ , $\vec{n}_i$ and $\vec{n}_j$ are normal vectors of adjacent points. If $\theta_{\text{diff}} < \theta_{\text{thres}}$ (we set $\theta_{\text{thres}} = 15^o$ ), $\vec{n}_i$ and $\vec{n}_j$ are determined that they belong to the same plane. This threshold ensures precise separation between the horizontal plane and the vertical plate. To avoid interference from segmented small planes, we remove clusters with fewer than 50 points (noise or small fragments) and retain the effective step plane.

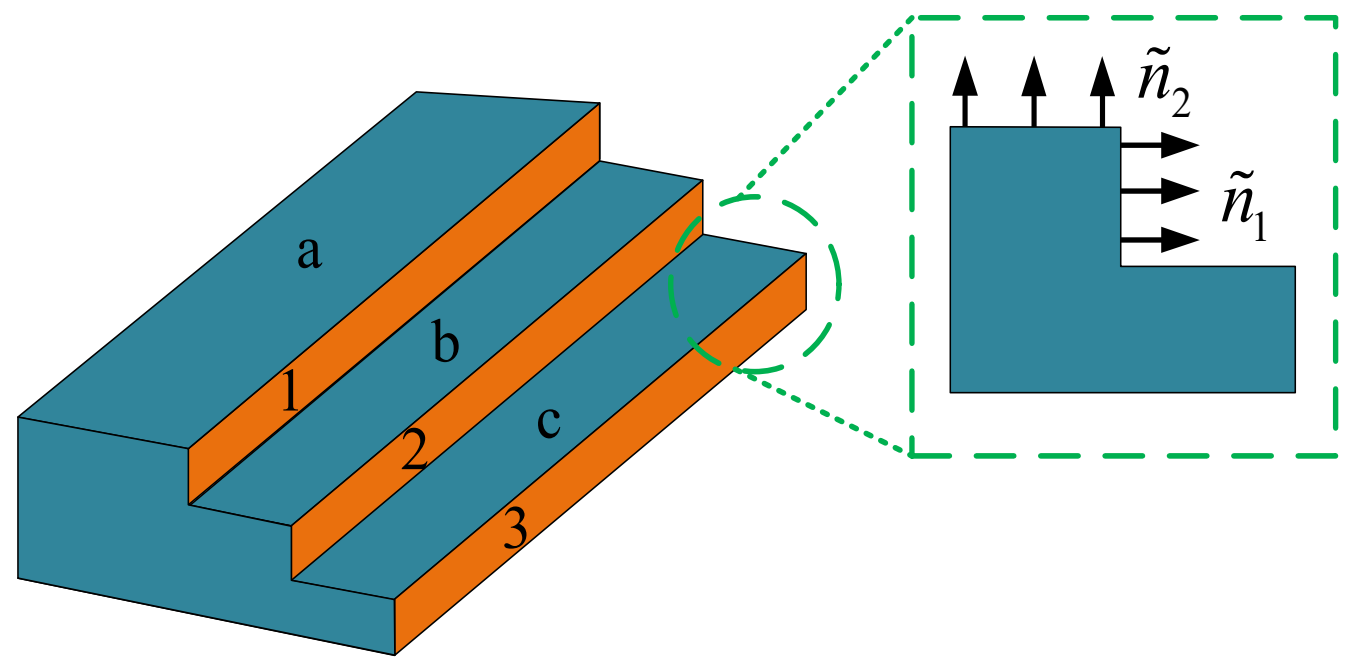

**Fig.4 Schematic diagram of staircase**

In our strategy, as shown in Algorithm 1, if the deep learning identifies that the results of the previous and next frames are different, the depth camera is triggered to collect the point cloud data of the scene. Since YOLOv11 has previously identified and obtained the ROI region of interest. Therefore, the processing and calculation of the point cloud can be performed in the local region of the ROI.

**Algorithm 1 Staircase scene recognition combined with point cloud**

**Input:** Front frame recognition result *F*, back frame recognition result *B* and point cloud segmetion result *P*.
**Output:** Accurate recognition results *I*.
1: **if** $B = F \wedge B \neq \varnothing \wedge F \neq \varnothing$ **then**
2: $I \leftarrow B$ ; // Step 1. Check whether the results of the frames front and back are consistent
3: **else if** $P = F$ **then**
4: $I \leftarrow F$ ; // Step 2. Due to the inconsistency between the results of front and back the frames, the point cloud segmentation result is consistent with the front frame, therefore the output result is the result of the front frame
5: **else**
6: $I \leftarrow B$ ;
7:**end if**
8:return $I$ ;

In order to speed up the algorithm and avoid the interference of non-target objects, Image processing calculations were performed on regions of interest previously obtained with YOLOv11. After the staircase point cloud is segmented in the region of interest, if the proportion of the total number of vertical plane point clouds $L_1$ and the total number of horizontal plane point clouds $L_2$ to the total number of point clouds $L$ in the region of interest exceeds threshold $T_1$, it is going upstairs. Similarly, if the proportion of the horizontal plane is greater than the set threshold $T_2$, it can be judged as going downstairs, as shown in Eq. (3). In this paper, $T_1$ and $T_2$ are 0.4 and 0.8, respectively, determined through statistical analysis of 100 diverse staircase samples (covering step heights of 15-20 cm and depths of 25-30 cm). For each sample, we computed the horizontal plane point cloud ratio across camera tilt angles from 0° to 30° (encompassing typical installation ranges), then fitted a logistic regression modelto optimize $T_2$=0.8 for >95% accuracy in downstairs identification. The lower threshold $T_1$=0.4 was set at the same way. At this point, the further detection is completed, and finally the detection of the terrain is completed.

$$\begin{cases} \frac{L_1}{L} > T_1, \frac{L_2}{L} > T_1 & \text{upstairs} \\ \frac{L_2}{L} > T_2 & \text{downstairs} \end{cases} \tag{3}$$

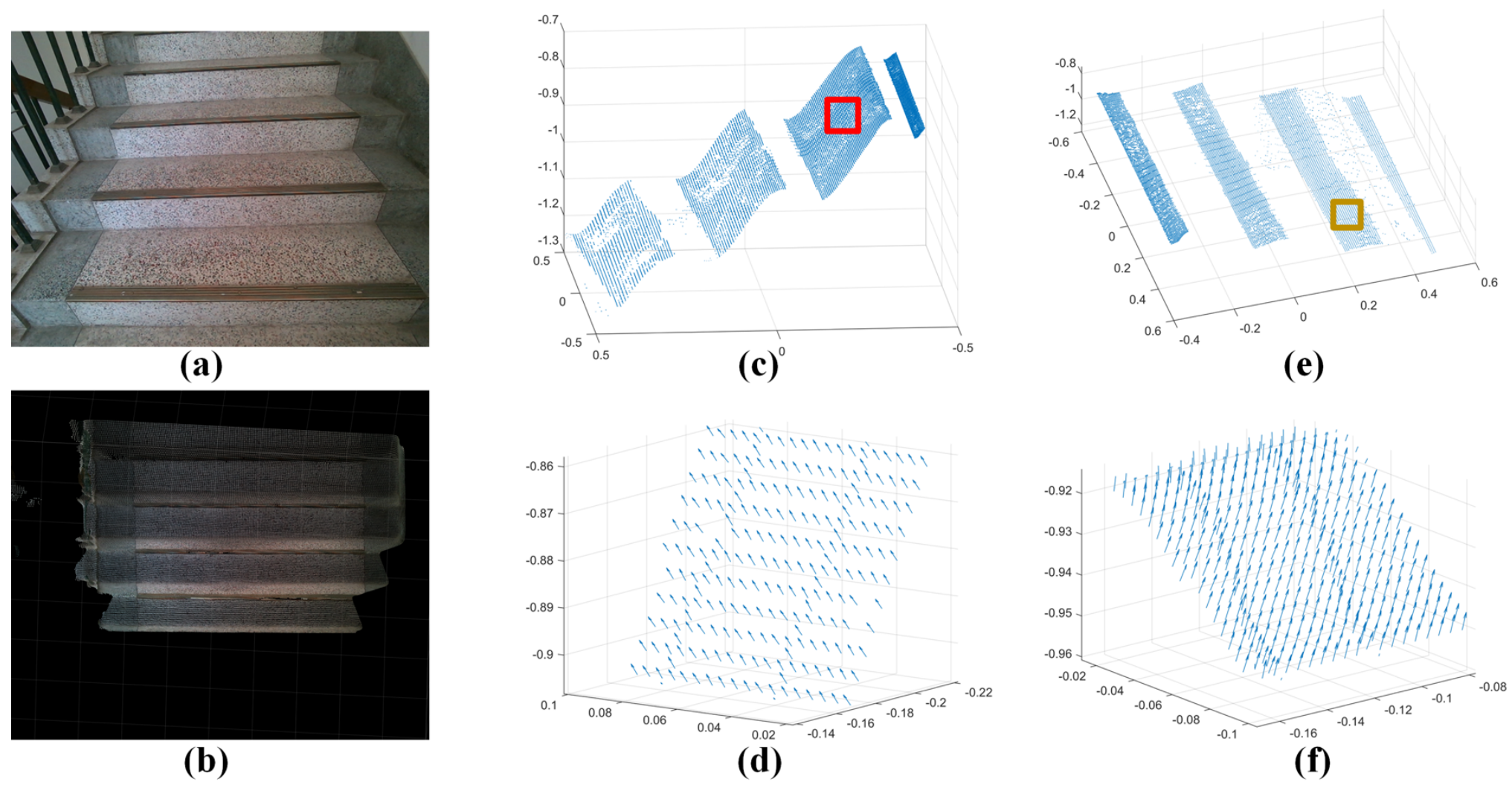

**Fig.5 Diagram of point cloud segmentation:(a) Monocular image; (b) Original point cloud; (c) Horizontal planes segmentation result; (d) Local enlarged view of the normal distribution in horizontal planes; (e) Vertical planes segmentation result; (f) Local enlarged view of the normal distribution in vertical planes**

Point cloud segmentation was tested 300 times on the point cloud obtained by the TOF camera. The plane is segmented according to the difference of the normal shown in Fig.5. The accuracy of this method can reach 96.7%. Therefore, this method is suitable for further detection. On the basis of the 97.1% recognition accuracy of the YOLOv11 model, the introduction of 3D depth information as redundant information can reduce false recognition and improve the recognition accuracy to 98.7%.

## III. Staircases reconstruction using binocular polarization and TOF fusion

In the aspect of 3D reconstruction of the scene, polarization 3D reconstruction is used in this section. This method relies on the polarization information of the target object, so polarization 3D reconstruction has low requirements for the reconstructed object and does not depend on the texture details of the object surface. Therefore, it can realize the 3D reconstruction of low-texture target objects and is less affected by illumination. However, it has the problem of the ambiguity of the polarization information. To solve this problem, here we introduce binocular polarization and TOF fusion to reach the high-quality reconstruction. In the process of the fusion, the calibration relationship between sensors is crucial.

### *A. Heterogeneous sensor calibration of monocular and TOF cameras*

#### *1. The Process of heterogeneous sensor calibration*

In the traditional dual target positioning algorithm, $P$ is a point in the common field of view of the left and right cameras. The ultimate goal of binocular calibration is to obtain the transfer matrix between the left and right cameras.

Before fusion, we calculate the relative pose between TOF and polarization camera. In this scheme, the resolution of the TOF camera is $640\times480$, and the resolution of the monocular polarization camera is $1224\times1024$, so the traditional binocular calibration method has limitations on this. The images captured by polarization and TOF camera are shown in Fig.6. To solve this problem, the heterogeneous sensor calibration method combines ICP registration and gray wolf optimization algorithm. The ICP registration algorithm is a 3D point cloud matching algorithm, assuming that there are two groups of point clouds to be registered, namely, the source point cloud $\boldsymbol{P}=\{\boldsymbol{p_1},\boldsymbol{p_2},\ldots,\boldsymbol{p_n}\}$ to be matched and the target point cloud $\boldsymbol{Q}=\{\boldsymbol{q_1},\boldsymbol{q_2},\ldots,\boldsymbol{q_n}\}$ to be finally matched. The transformation matrix between the two sets of point clouds is calculated by the registration algorithm, and the ultimate goal is to minimize the error function after registration. The error function can be expressed as:

$$E(\boldsymbol{R},\boldsymbol{T})=\frac{1}{n}\sum_{i=1}^{n}\left\|\boldsymbol{q}_i-(\boldsymbol{R}\boldsymbol{p}_i+\boldsymbol{T})\right\|^2 \tag{4}$$

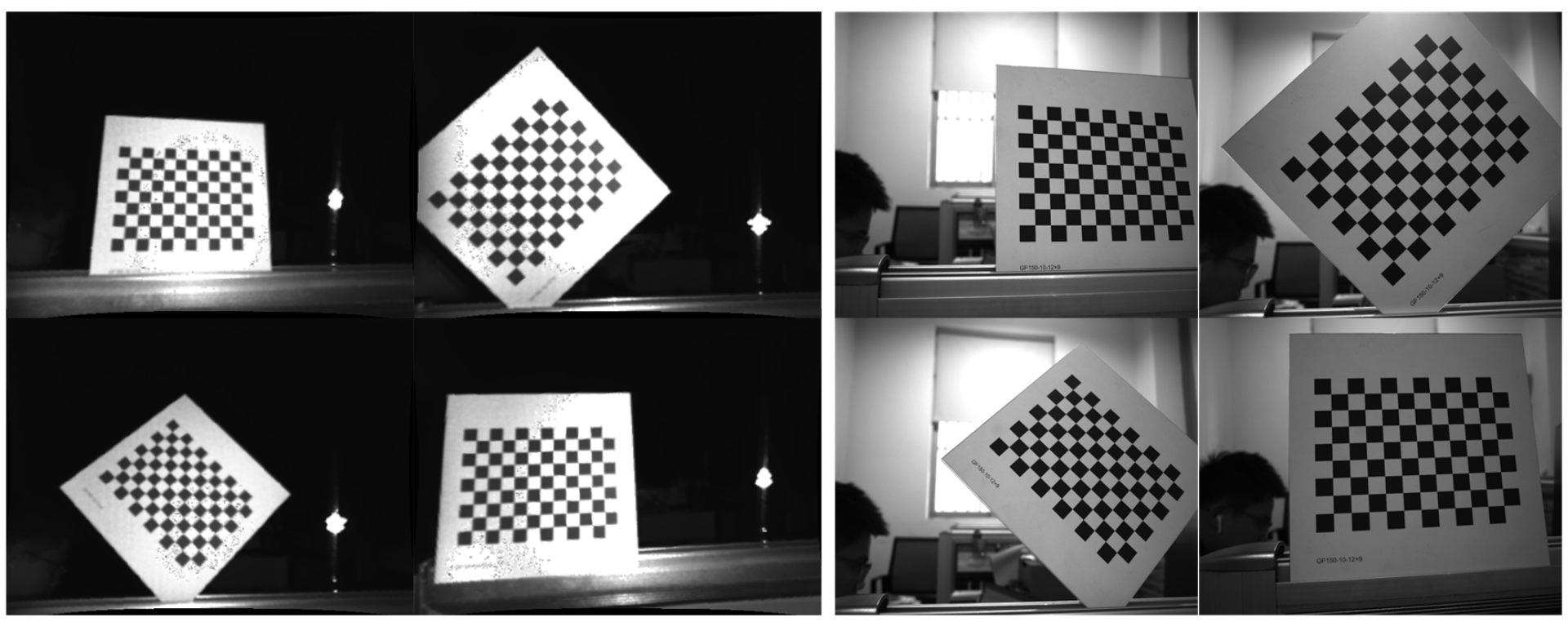

**(a)** **(b)**

**Fig.6 Part of the images of calibration board: (a) Images captured by TOF camera; (b) Images captured by polarization camera**

In order to avoid the inaccurate results caused by the ICP algorithm falling into the local minimum or the errors of the two original point clouds, which ultimately affect the registration results between the source point cloud and the target point cloud, the checkerboard calibration board is used in this paper to combine the ICP registration and the gray wolf optimization algorithm for heterogeneous sensor calibration. The steps of heterogeneous sensor calibration are as follows:

(1) a polarization camera and a TOF camera are rigidly and fixedly connected, a calibration plate is placed in the common visual field of the polarization camera and the TOF camera, images of the calibration plate are collected at the same time, and 20 groups of images are taken;

(2) According to the internal reference data of the camera and the world coordinate system of the corner point of the calibration board, the pose of the calibration board can be estimated to obtain the rotation and translation matrix of the calibration board under the camera coordinate system, and the world coordinate of the chessboard corner point is converted and calculated to obtain the 3D point cloud data of the corner point under the camera coordinate system. The point cloud obtained by the TOF camera is $\boldsymbol{P}=\{\boldsymbol{p_1},\boldsymbol{p_2},\ldots,\boldsymbol{p_n}\}$, the point cloud obtained by the polarization camera is $\boldsymbol{Q}=\{\boldsymbol{q_1},\boldsymbol{q_2},\ldots,\boldsymbol{q_n}\}$. The initial transformation rotation matrix $\boldsymbol{R}$ and the translation vector $\boldsymbol{T}$ of the relative pose relationship between the polarization camera and the TOF camera are calculated by Zhang's calibration method for the ICP registration algorithm [35], ensuring that the algorithm is optimized from a high-quality starting point to avoid the uncertainty of random initial values.

(3) Based on the relative pose relationship between the polarization camera and the TOF camera obtained from the dual objective method, the initial values for ICP registration are iteratively solved to obtain the relatively accurate pose relationship between point clouds $\boldsymbol{P}$ and $\boldsymbol{Q}$;

(4) In order to avoid the defect of ICP algorithm easily falling into local optima and ultimately leading to inaccurate matching results, the grey wolf optimization algorithm is used to further optimize the fitness function of Eq. (4) for the matching results. Due to its hierarchical leadership mechanism, it avoids the convergence of a single guiding source compared to other optimization algorithms.

*2. Improved grey wolf optimization algorithm*

The grey wolf optimization algorithm is a bionic optimization algorithm, which refers to the hierarchical system and some habits of hunting behavior in the grey wolf population, and has the advantages of simple parameters and easy implementation [36]. In comparison, particle swarm optimization algorithm is prone to getting stuck in local optima in high-dimensional nonlinear parameter optimization problems involving 6-degree-of-freedom rotation matrices and translation vectors due to the homogenization of particle behavior. grey wolf optimization has introduced a hierarchical leadership mechanism that better meets the requirement of "global coarse adjustment first, followed by local fine adjustment" in calibration problems. And the inertia weight of particle swarm optimization needs to be manually set, usually fixed or linearly attenuated, which makes it difficult to adaptively adjust the search intensity. This paper improves grey wolf optimization through dynamic weight allocation. We input the rotation matrix $\boldsymbol{R}$ and translation vector $\boldsymbol{T}$ obtained from ICP registration as initial values to the grey wolf optimization algorithm, and optimize these two parameters as the final goal. The optimization process of the algorithm can be divided into two stages. In the early stage, the core task of global search is to initially lock the region where the global optimal solution is most likely to appear. In order to achieve this goal, we need to ensure that the search range covers the whole search space and prevent any potential optimal solution candidates from being missed, so as to locate the optimal solution more accurately. In the later stage of local search, it focuses on carrying out small-scale fine search near the location determined by global search. This step aims to further refine the position of the optimal solution obtained by global search, so as to obtain a more accurate solution.

The grey wolf population has a strict hierarchy, and the behavior of the population is also reflected in the gray wolf optimization algorithm. Suppose there are three grey wolves (*α, β, δ*) represent in the wolf pack. Among them, wolf $\alpha$ has a higher level of status, which determines the behavior of the wolf pack to the greatest extent. In each position update process, the grey wolf position updated in the previous update will be used as a reference. In the d-dimensional search space, the position of the *k-th* grey wolf individual at the *t-th* iteration is denoted as $X_k(t)$, and this position update mechanism helps the grey wolf group to gradually approach the optimal solution in the search space. The mathematical model of grey wolf tracking prey can be expressed as:

$$D = \left| CX_p(t) - X_k(t) \right| \tag{5}$$

$$X_k(t+1) = X_p(t) - AD \tag{6}$$

where Eq. (5) is the expression for the search step, $X_p(t)$ represents the current position of the prey, and $X_k(t+1)$ represents the updated position of the current individual. In addition, $A$ and $C$ play a key role in the algorithm as disturbance factors, which are defined as follows:

$$A = 2a \cdot rand(\ ) - a \tag{7}$$

$$C = 2 \cdot rand(\ ) \tag{8}$$

where $rand(\ )$ is a random number between [0,1], $a$ is the convergence factor defined as:

$$a = 2\times(1-\frac{t}{T}) \tag{9}$$

where, $T$ is the total number of iterations. The position update formula of wolf $\alpha$, wolf $\beta$ and wolf $\delta$ of the population individuals is:

$$\begin{cases} D_\alpha = \left|C_1 X_\alpha(t) - X_k(t)\right| \\ D_\beta = \left|C_2 X_\beta(t) - X_k(t)\right| \\ D_\delta = \left|C_3 X_\delta(t) - X_k(t)\right| \end{cases} \tag{10}$$

$$\begin{cases} X_1(t+1) = X_\alpha(t) - A_1 D_\alpha \\ X_2(t+1) = X_\beta(t) - A_2 D_\beta \\ X_3(t+1) = X_\delta(t) - A_3 D_\delta \end{cases} \tag{11}$$

$$X_k(t+1) = \frac{1}{3}\sum_{i=1,2,3} X_i(t+1) \tag{12}$$

where, $X_\alpha(t)$, $X_\beta(t)$ and $X_\delta(t)$ respectively represent the current position of wolf $\alpha$, wolf $\beta$ and wolf $\delta$. $D_\alpha$, $D_\beta$ and $D_\delta$ correspond to the search steps of the three-headed wolf, and they determine the movement distance of different wolf in the search process. $X_i(t+1)(i=1,2,3)$ represents the position of the individual grey wolf after the guidance update of wolf $\alpha$, wolf $\beta$ and wolf $\delta$. $A_i(i=1,2,3)$ and $C_i(t+1)(i=1,2,3)$ are the perturbation factors of Eq. (7)-(8).

In the grey wolf optimization algorithm, it can be seen from Eq. (12) that the position of the next movement of the population needs to be jointly decided by wolf $\alpha$, wolf $\beta$ and wolf $\delta$.All three wolves equally contribute to the same in updating the position of the movement, but this decision-making method cannot reflect the guidance and importance of the three wolves to the decision-making of the group leaders. In order to realize the self-adaptive adjustment of the weight proportion of the decision-making individual in different periods, in each iterative calculation process of the algorithm, the dynamic weight is added to the Eq. (12) to make the update position of the population more reliable. The improved update position of the population is:

$$X_k(t+1) = \frac{(X_1(t+1))^2 + (X_2(t+1))^2 + (X_3(t+1))^2}{3\sum_{i=1,2,3} X_i(t+1)} \quad (i=1,2,3) \tag{13}$$

By dynamically allocating weights, the convergence of the group towards the global optimal solution is accelerated, reducing the risk of local optima caused by initial ICP value deviation. In the Grey Wolf Optimization algorithm, $\alpha$, $\beta$, and $\delta$ wolves represent the current best, second-best, and third-best solutions in the population, respectively. The position of the α wolf is closest to the global optimum (or the region where the global optimum is located). Dynamic weight adjustment (such as enhancing the contribution of the α wolf through a squared term) essentially allows the population to "trust" more in the guidance of high-quality solutions during updates. High-quality solutions contain richer information about the global optimum, and the search direction they guide is more likely to point towards the global optimum rather than local optimum regions. After enhancing the weight of high-quality solutions, in each iteration, individuals approach the high-quality solutions with a larger magnitude, making the population's approximation path towards the global optimum shorter and more direct within the same number of iterations.

In the position update of Eq. (13) based on weight, this paper introduces the strategy of Levy flight method. Levy flight is a random walking pattern with non-Gaussian characteristics, and its step size distribution exhibits an alternating pattern of short distance exploration and long-distance jumping, which is significantly different from the uniform step size of traditional random search. The ICP algorithm achieves registration by iteratively minimizing the point cloud matching error Eq. (4), but it relies on initial values and is prone to being trapped in local optima (such as the minimum error caused by calibration plate corner noise). Levy's long-distance jumps during flight enable the grey wolf optimization algorithm to escape from the local optimal region locked by ICP and search for better solutions in a broader parameter space (possible values of rotation matrix ***R*** and displacement vector ***T***). $X_k^{'}(t+1)$ is the grey wolf individual position updated by the Levy flight method:

$$X_k^{'}(t+1) = L_v(X_\alpha(t) - X_k(t)) + X_k(t) \tag{14}$$

where, $X_k^{'}(t+1)$ superimposes a random disturbance with Levy distribution characteristics on the basis of $X_k^{'}(t+1)$, which helps to escape from the local optimal region and explore a broader solution space., and $L_v$ is the flight step length, which is defined as:

$$L_v = \frac{u}{|v|^{1/\beta}} \tag{15}$$

where, $u$ and $v$ are normally distributed. $u \sim N(0,\sigma_u^2)$ and $v \sim N(0,1)$ are independent normally distributed random variables. $\sigma_u$ is a parameter in the calculation of Levy flight step size, and its physical meaning is related to the characteristics of Levy distribution, which is used to describe the normal distribution standard deviation of the random variable *u* and defined as:

$$\sigma_u = \frac{\Gamma(1+\beta)\sin(\beta\pi/2)}{2^{(\beta-1)/2}\Gamma((1+\beta)/2)\beta} \tag{16}$$

where, $\beta$ is usually taken in the range [0,2] and is taken as 1.5 in this paper, so $\sigma_u = 0.58$.

In order to ensure the search efficiency in the iterative process, the principle of greedy algorithm is used to compare the fitness values of Eq. (13) and (14), and the individual with better fitness is used as the final update position, which minimizes the error in Eq. (4):

$$X_k(t+1) = \begin{cases} X_k(t+1) & \text{if} f(X_k(t+1)) < f(X_{k'}(t+1)) \\ X_{k'}(t+1) & \text{otherwise} \end{cases} \tag{17}$$

where, $f(\cdot)$ is the fitness function (i.e., the error function in Eq. (4)), ensuring that the algorithm strikes a balance between local fine search (dynamic weighting) and global exploration (Levy flight), preserving the local accuracy of ICP while enhancing the robustness of global registration. Finally, this algorithm converges at 40th iteration.

*B. Polarization gradient field ambiguity correction with binocular and TOF fusion*

In the process of polarization imaging, more dimensional information can be obtained compared with the ordinary image, and the zenith angle $\theta$ and azimuth angle $\phi$ of the surface can be obtained through the polarization information in the collected polarization image, and then the normal vector parameters of the surface can be obtained. Therefore, the reconstruction result of polarization 3D reconstruction is not related to the amount of texture on the surface of the object, and has certain advantages for low-texture objects, and can effectively avoid the disadvantage that binocular 3D reconstruction depends on texture information. In polarized vision, as shown in Fig.7, zenith angle $\theta$ is the angle between the surface normal of the target object and the centerline of the camera optical axis, and azimuth angle $\phi$ is the angle between the surface normal ***N*** of the target object projected onto the image coordinate system plane and the clockwise direction of the positive x-axis of the image coordinate system. The gradient field is composed of surface normal calculated from zenith and azimuth angles.

Common staircase materials such as concrete, wood, stone, etc. have relatively rough surfaces and many uneven structures at the microscopic level. According to the principle of reflection, when light shines on these rough surfaces, it will scatter in all directions, forming diffuse reflection. In practical environments, staircases are often illuminated by light from multiple directions, such as natural light entering through windows or indoor lighting shining from different angles. These nonunidirectional light rays will generate complex reflections on the surface of the staircase, further increasing the proportion of diffuse reflections. Because the characteristic of diffuse reflection is the ability to uniformly reflect light at different angles and directions, diffuse reflection is easier to observe and capture under the illumination of multi-directional light. The imaging devices commonly used to obtain staircase images, such as cameras, have relatively fixed viewing angles, making it difficult to capture a specific angle that can capture a large amount of specular reflections. And diffuse reflection can be observed well from various angles, so the camera is more likely to capture the image information formed by diffuse reflection. Even though some specular reflections may be observed at certain specific angles, due to the complexity and unevenness of the staircase surface, these specular reflections are often local and incomplete, and their proportion in the entire image is still relatively small compared to large-area diffuse reflections.

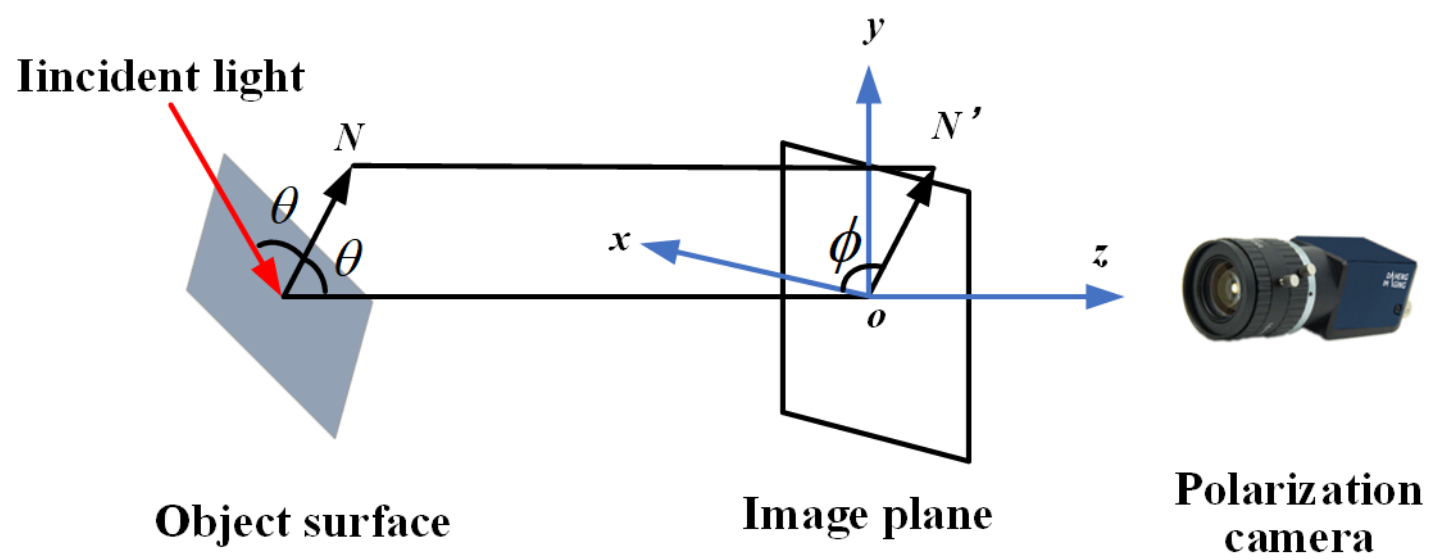

**Fig.7 Schematic diagram of polarization zenith angle and azimuth angle**

In this paper, we assume reflected light is primarily diffuse. As shown in Fig.8, the diffuse reflection light on the surface of an object is the component of light that enters the interior of the object through the surface and is scattered by internal particles before being refracted back into the air. The azimuth angle $\phi$ is:

$$\phi = \frac{1}{2}\tan^{-1}\left(\frac{s_2}{s_1}\right) \tag{18}$$

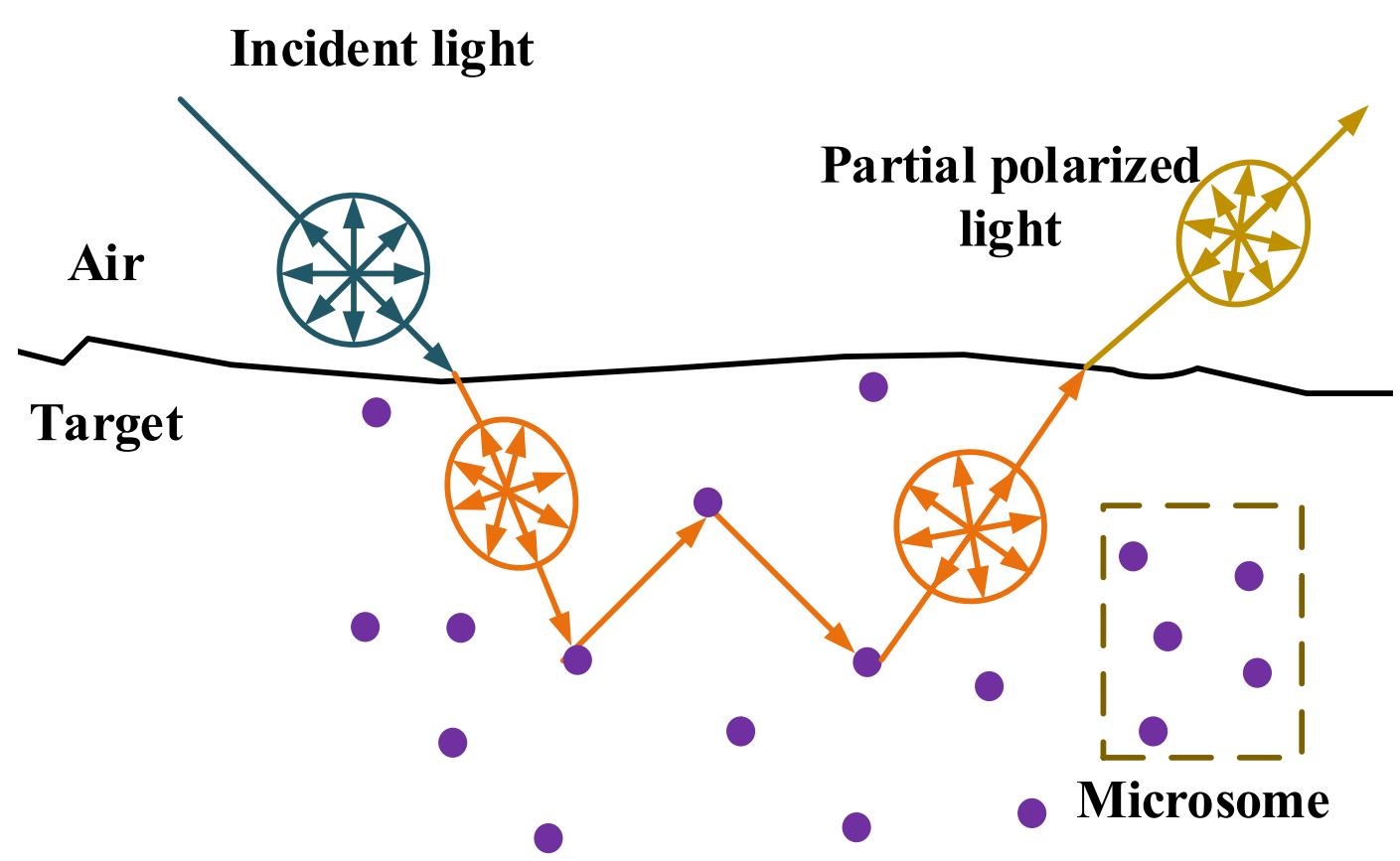

**Fig.8 Schematic diagram of diffuse reflection**

Another method of obtaining azimuth angle $\phi$ is to adjust the angle of the polarizer by rotating it. The default polarization angle $\varphi$ at which the light intensity value reaches its peak for the first time during rotation is considered as the azimuth angle $\phi$. Due to the interval range of azimuth angle $\phi$ being $[0, 2\pi]$ and the interval range of polarization angle being $[0, \pi]$, there is ambiguity in the solution of azimuth angle $\phi$, and the result of azimuth angle is $\phi = \varphi$ or $\phi = \varphi + \pi$.

Due to the manual operation required for the rotation process of the polarizer, there are certain errors in the rotating mechanical structure and experimental equipment. Therefore, the polarization images in this paper were obtained through a polarization camera. Combining Fresnel theory, the functions of DoP $\rho$ and zenith angle $\theta$ can be determined [32-35]:

$$\rho = \frac{(n - 1/n)^2 \sin^2\theta}{2 + 2n^2 - (n + 1/n)^2 \sin^2\theta + 4\cos\theta\sqrt{n^2 - \sin^2\theta}} \tag{19}$$

where $n$ represents the reflection coefficient. The exact value of zenith angle can be obtained by solving Eq. (19). As shown in Fig.9, Eq. (19) is a monotonic function, therefore, there is no zenith angle ambiguity in the diffuse reflection model. Based on the above analysis, in the process of 3D reconstruction of diffuse reflection polarization, there are the problems of polarization azimuth ambiguity and the noise of polarization gradient field in the case of low zenith angle, so binocular stereo vision is introduced into the 3D reconstruction of polarization to correct the polarization gradient field preliminarily. Then the 3D information of TOF is used for local correction. The specific correction scheme is shown in Fig.10.

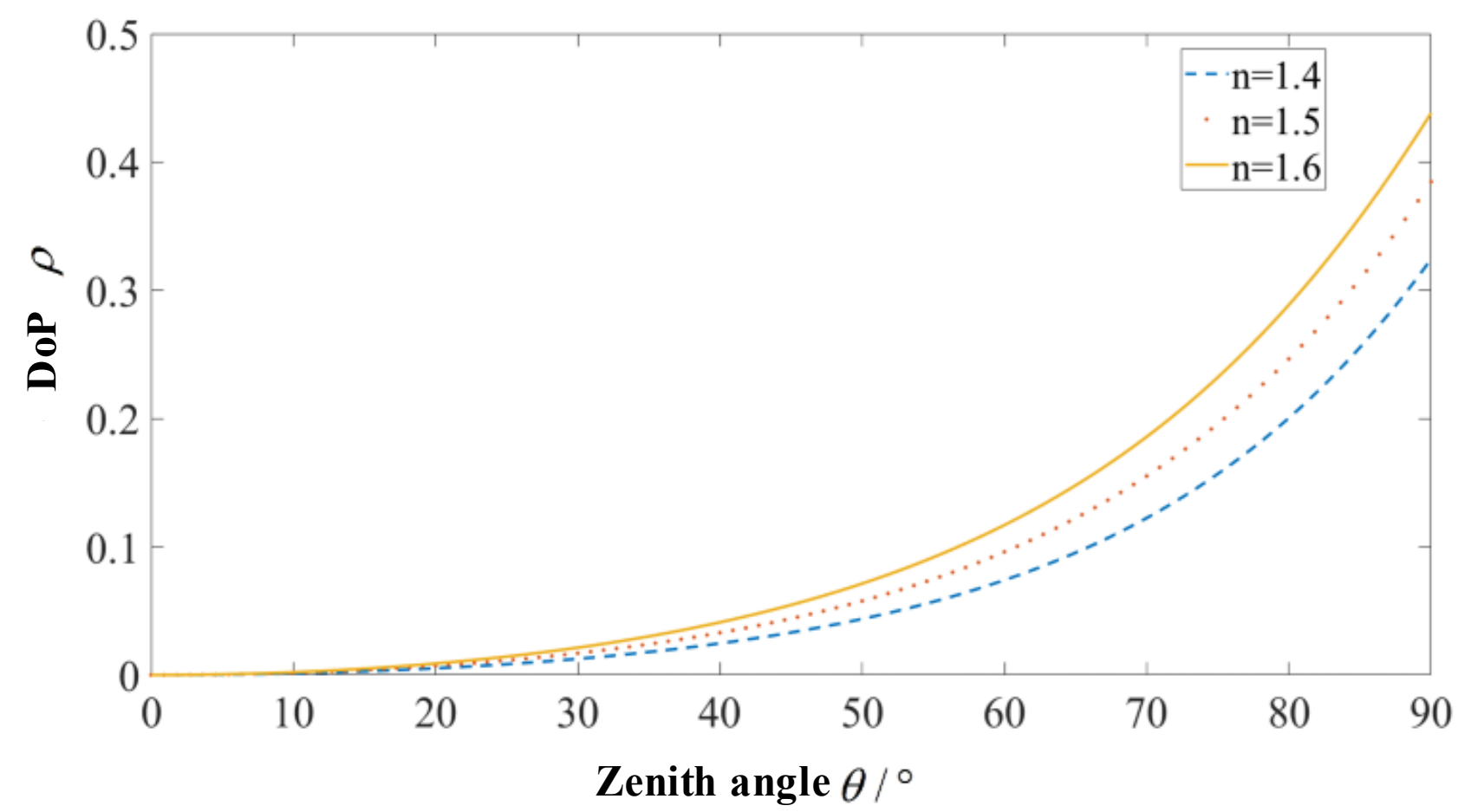

**Fig.9 Schematic diagram of the relationship between DoP and zenith angle in diffuse reflection model**

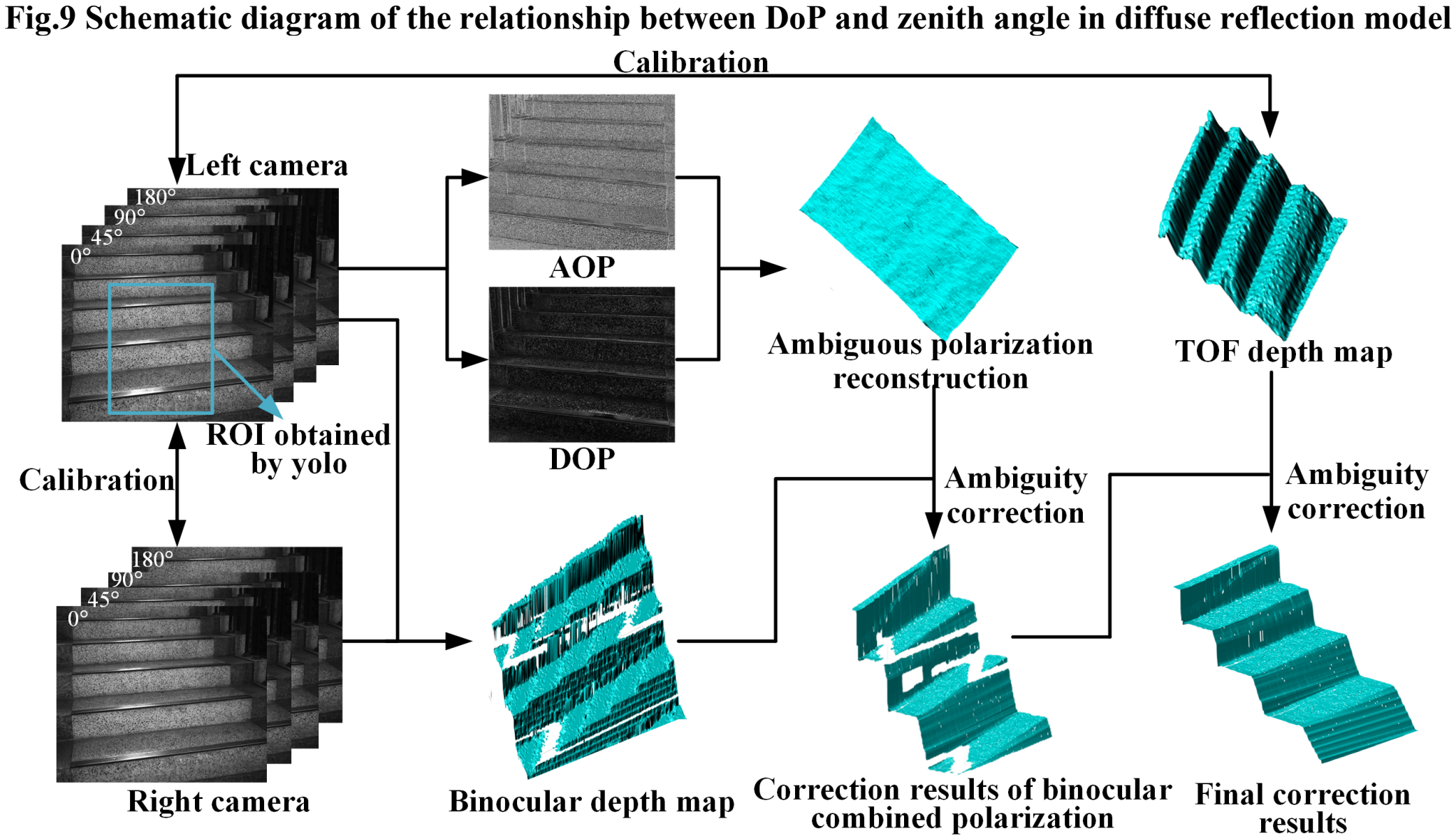

**Fig.10 Polarization gradient field ambiguity correction process**

The first stage is the acquisition of binocular depth images of the ROI obtained by staircase recognition. The left and right images are captured by binocular polarization camera, and then the binocular depth map is obtained by binocular imaging method. The second stage is the acquisition of the polarization depth image, zenith and azimuth angles are derived from the polarization image acquired by the polarization camera, the polarization gradient field is computed based on the zenith angle and the azimuth angle, the azimuth angle ambiguity of the gradient field is corrected through the binocular depth image, and then the surface relative height of the gradient field is calculated by using the integral method. In the third stage, the normal vector of the polarization gradient field in the hole area of the binocular depth map is corrected by the normal vector of the depth map collected by the TOF camera to correct the ambiguity of the local polarization gradient field.

The zenith angle $\theta$ and azimuth angle $\varphi$ of the object surface can be obtained according to the polarization imaging, so the normal vector $\boldsymbol{N}^{\boldsymbol{P}} = \left[\boldsymbol{N}_x^{\boldsymbol{P}}, \boldsymbol{N}_y^{\boldsymbol{P}}, \boldsymbol{N}_z^{\boldsymbol{P}}\right] = \left[\tan\theta\cos\phi, \tan\theta\sin\phi, 1\right]$ of the object surface can be obtained according to the polarization characteristics:

Obtaining the surface normal $\boldsymbol{N}^{\boldsymbol{B}} = \left[\boldsymbol{N}_x^{\boldsymbol{B}}, \boldsymbol{N}_y^{\boldsymbol{B}}, \boldsymbol{N}_z^{\boldsymbol{B}}\right]$ of the binocular depth image through the difference approximate derivative.

Comparing the polarization normal $\boldsymbol{N}^{\boldsymbol{P}}$ with the binocular normal $\boldsymbol{N}^{\boldsymbol{B}}$. $\boldsymbol{N}^{\boldsymbol{P}}$ will become the negative value of the true normal vector due to the 180 ° ambiguity of the azimuth angle, while the binocular normal $\boldsymbol{N}^{\boldsymbol{B}}$ is obtained from the rough depth map through the differential approximate derivative, and the direction is correct. Therefore, the polarization normal vector $\boldsymbol{N}^{\boldsymbol{P}}$ can be corrected by the binocular normal vector $\boldsymbol{N}^{\boldsymbol{B}}$. The optimization goal $\Lambda^*$ can be established as:

$$\Lambda^* = \arg\min \sum_{i \in R_1} [(\boldsymbol{N}_{xi}^{\boldsymbol{P}} - \Lambda_i \bullet \boldsymbol{N}_{xi}^{\boldsymbol{B}})^2 + (\boldsymbol{N}_{yi}^{\boldsymbol{P}} - \Lambda_i \bullet \boldsymbol{N}_{yi}^{\boldsymbol{B}})^2]^{1/2},\ \Lambda_i \in \{-1, 1\} \tag{20}$$

where, $R_1$ represents the binocular normal vector region.

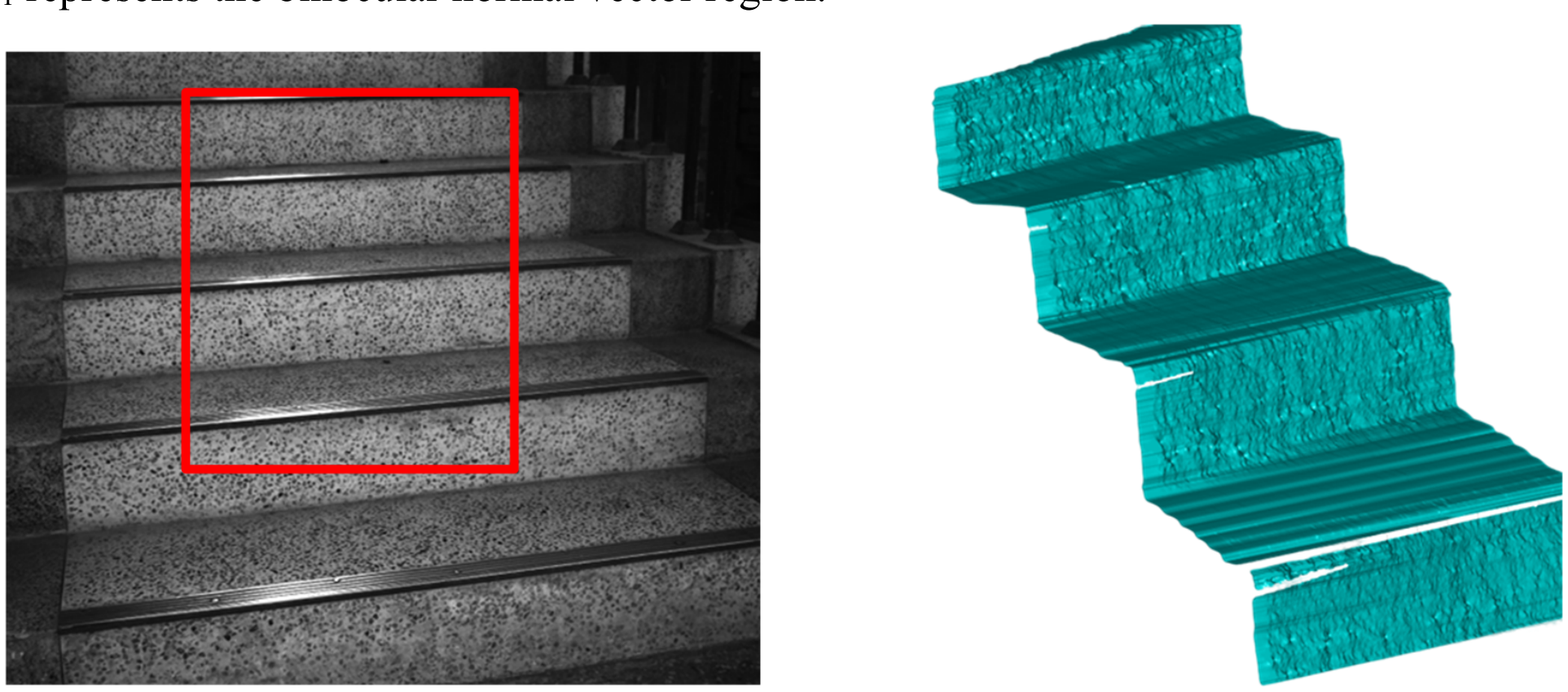

**(a)** **(b)**

**Fig.11 Polarization Gradient Field Integration Results: (a) Original image; (b) Integral results**

Although the polarization gradient field is corrected for azimuth ambiguity by binocular depth information, there will be some low-frequency information missing in the binocular imaging process, which will lead to the binocular depth map hole. For the uncorrected normal vector corresponding to the hole region, the fusion TOF method is introduced in this paper. Firstly, the 3D information acquired by the TOF is converted into an extrinsic parameter matrix through heterogeneous sensor calibration , and the reference coordinate system of the extrinsic parameter matrix is set as the camera coordinate system of the left camera, and the extrinsic parameter matrix is converted into a depth map, and the normal vector $\boldsymbol{N}^{\boldsymbol{T}}$ corresponding to the depth map is obtained through derivation, and the direction of the normal vector is correct. Secondly, the hole region in the binocular depth map is obtained through the Canny algorithm, and the normal vector corresponding to the area is extracted from the TOF depth map. Therefore, the polarization normal vector $\boldsymbol{N}^{Ph}$ in the hole region can be corrected by the TOF normal vector $\boldsymbol{N}^{T}$. The optimization goal $\Lambda^{**}$ can be established as:

$$\Lambda^{**} = \arg\min \sum_{i \in R_1} [(\boldsymbol{N}_{xi}^{Ph} - \Lambda_i \bullet \boldsymbol{N}_{xi}^{T})^2 + (\boldsymbol{N}_{yi}^{Ph} - \Lambda_i \bullet \boldsymbol{N}_{yi}^{T})^2]^{1/2},\ \Lambda_i \in \{-1,1\} \tag{21}$$

Thus, the azimuth correction of the staircase polarization information is completed. The polarization gradient field constructed by the corrected polarization normal is processed by the integral algorithm [37] to obtain the relative height of the polarization surface as shown in Fig.11.

## IV. Experiment

*A. Experimental equipment*

The training platform used by the scene recognition algorithm is configured with Intel I5 12490F CPU, RTX 3060 GPU with 12G graphics memory, 32G storage capacity and DCAM550 TOF camera from Vzense Technology. Software used for deep learning includes Pycharm, Anaconda, and Pytorch.

The experimental platform for 3D reconstruction of polarization used in this paper consists of polarization camera, TOF camera, lens and camera holder, all of which are shown in Fig.12. The polarization camera is Daheng MER-502-79U3M-L, the resolution of the polarization angle image collected by the camera is 1024 × 1224. The camera is equipped with IMX250 MZR CMOS chip and its exposure mode is global exposure. A Hikrobot MF0828M-8 MP lens is used. The TOF camera and the TOF camera used in the scene recognition algorithm are of the same model, both of which are DCAM550 produced by Vzense Technology. The camera bracket is a self-designed rigid device to fix the relative position between the polarization camera and the TOF camera. The length of the binocular baseline constructed by the polarization camera is 50mm, and the total weight of the verification platform is 549g. In order to ensure the synchronization of data acquisition between polarization camera and TOF camera, we adopted a hardware external triggering mechanism.

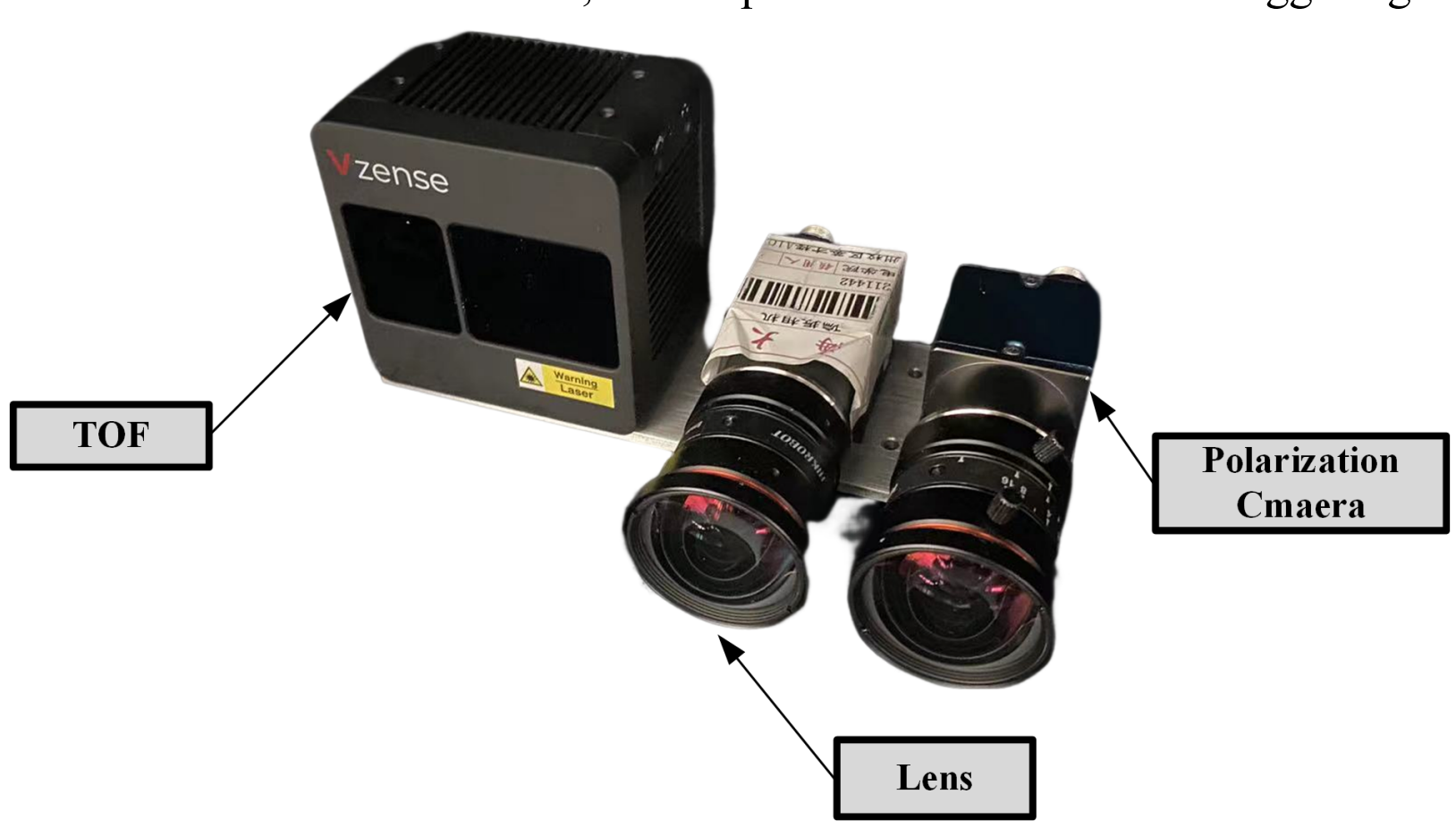

**Fig.12 experimental platform for 3D reconstruction**

*B. Robot staircase recognition experiment*

Dataset is the basis of deep learning model training. The current staircase dataset is ExoNet dataset from Laschowski team in Canada [38]. In order to ensure that the dataset conforms to the architectural style of our country and adapts to the recognition of domestic staircase, it is necessary to collect proprietary image data. Therefore, on the basis of the existing part of the data set, some self-collected staircase images are added as a supplement. The collection areas are from streets, shopping malls, schools, etc., including images of stairs with different lighting conditions, foreign body overlays, and different architectural styles. A subset of the images in the dataset is shown in Fig.13. The final training data set contains 16453 images containing the scene of staircases. The training results of the above models are shown in TABLE I. To enhance the robustness analysis of our method, the entire dataset was randomly divided into 5 mutually exclusive subsets (each subset containing 3291 images). We trained the model using 4 subsets for each experiment, and test the remaining 1 subset. This process is repeated 5 times, testing a different subset each time. The accuracy rates of the five experiments are 98.63%, 98.71%, 98.76%, 98.60%, and 98.80%, respectively. The average accuracy rate is 98.7%, the standard deviation is 0.08%, and the 95% confidence interval is 98.7% ± 0.10%.

The comparison in TABLE I. proves the feasibility of our improvement on the network. The recognition performance of the improved recognition algorithm fused with 3D information is significantly improved. Compared with YOLOv11, the accuracy and frame rate are increased by 3.7%. The final recognition result is shown in Fig.14. It can be seen that the algorithm proposed in this paper can accurately generate the bounding box of the target object. And the confidence evaluation of the scene recognition results is given.

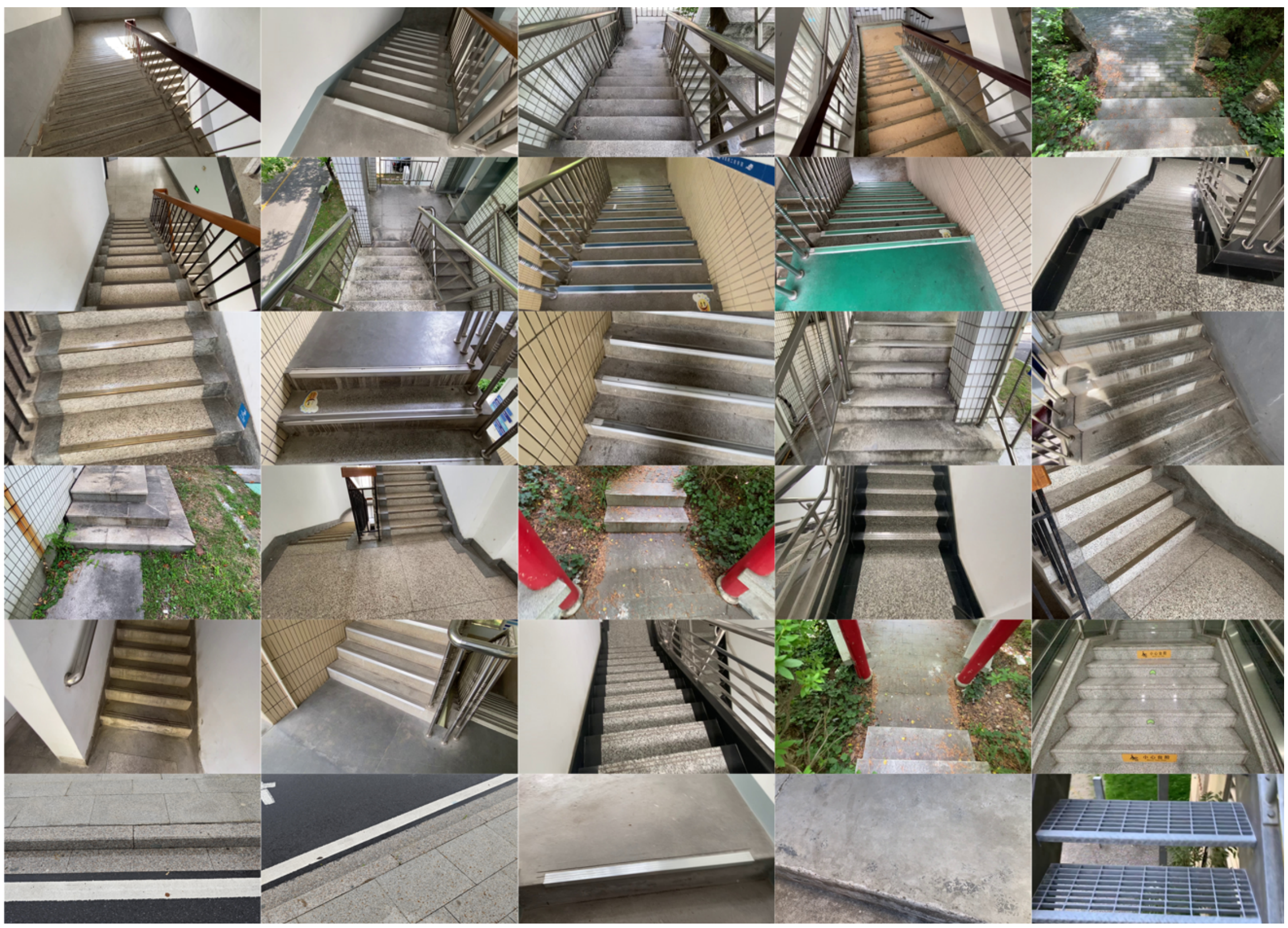

**Fig.13. Schematic diagram of the images in part of the datasets**

TABLE I COMPARISON OF RECOGNITION ALGORITHM RESULTS

| Model | mAP/(%) |
|---|---|
| YOLOv5s | 96.2% |
| YOLOv5-Lite | 92.5% |
| YOLOv7 [39] | 96.3% |
| SSD [40] | 98.2% |
| Faster-RCNN [41] | 87.8% |
| StairNetV3 [42] | 82.3% |
| Point cloud segmentation | 96.7% |
| YOLOv11 | 97.1% |
| The algorithm proposed in this paper | 98.7% |

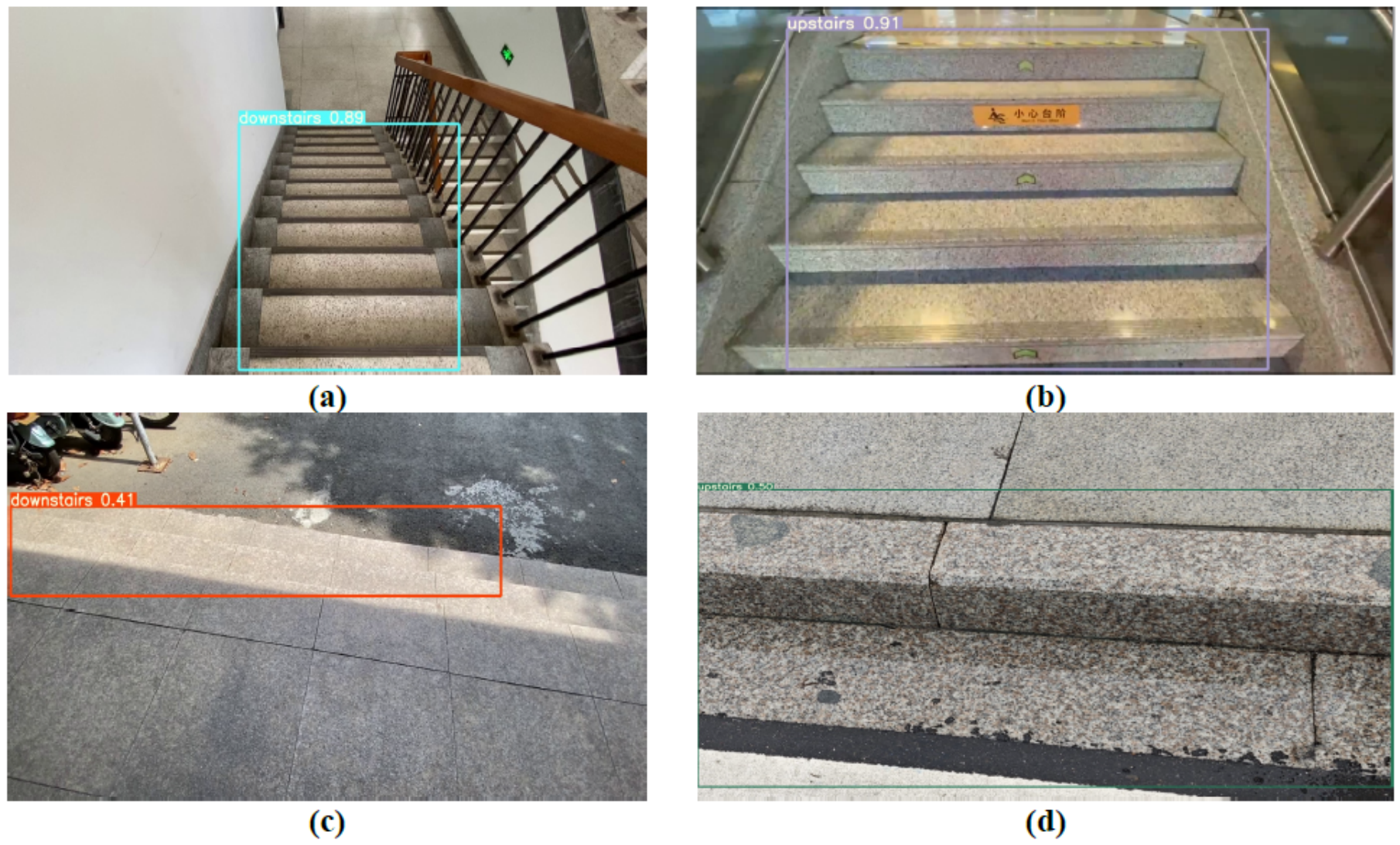

**Fig.14 Scene recognition results: (a)(c) Downstairs; (b)(d) Upstairs**

*C. Experimental of heterogeneous sensor calibration*

The heterogeneous sensor calibration method for ICP registration based on the improved grey wolf optimization algorithm can reduce the mismatching generated during the registration of the ICP algorithm. The corner points of the calibration plates under the public view of the two cameras have a corresponding relationship in the point cloud without the interference of the point cloud noise. The improved grey wolf optimization algorithm introduced to the ICP algorithm further improves the calibration precision. The checkerboard selected in this paper is a 12×9 calibration plate with a unit length of 20 mm captured at the distance of 0.5m, 0.8m and 1m for total 21 items paired left-right images.

Firstly, the internal parameter calibration of the polarization camera and the TOF camera is completed by using the traditional calibration method to obtain the internal parameter matrix and the radial distortion coefficient of the camera, and the internal parameter matrix of the TOF camera is:

$$\boldsymbol{K}_p = \begin{bmatrix} 456.4448 & 0 & 336.2882 \\ 0 & 457.1441 & 252.9748 \\ 0 & 0 & 1 \end{bmatrix} \tag{21}$$

The intrinsic matrix of the polarization camera is:

$$\boldsymbol{K}_q = \begin{bmatrix} 1846.2992 & 0 & 604.0391 \\ 0 & 1846.9653 & 518.9741 \\ 0 & 0 & 1 \end{bmatrix} \tag{22}$$

Using the heterogeneous sensor calibration algorithm proposed in this paper, the rigid transformation matrix between the polarization camera and the TOF camera is calculated as:

$$\left[\boldsymbol{R},\boldsymbol{T}\right] = \begin{bmatrix} 1 & 0.0007 & 0.007 & -91.0947 \\ -0.0007 & 1 & -0.0016 & 16.5543 \\ -0.007 & 0.0016 & 1 & -22.8737 \end{bmatrix} \tag{23}$$

$\left[\boldsymbol{R},\boldsymbol{T}\right]$ is the relative pose relationship between the TOF camera and the polarization camera, and the schematic diagram of the relative pose relationship between the collected angular point cloud and the camera is shown in Fig.15. The result of the registration using the rigid transformation matrix between the polarization camera and the TOF camera is shown in Fig.16. By calculating the reprojection error for 10 times, the average calibration accuracy of this algorithm is 0.33±0.04 mm and the standard deviation is 0.12mm.

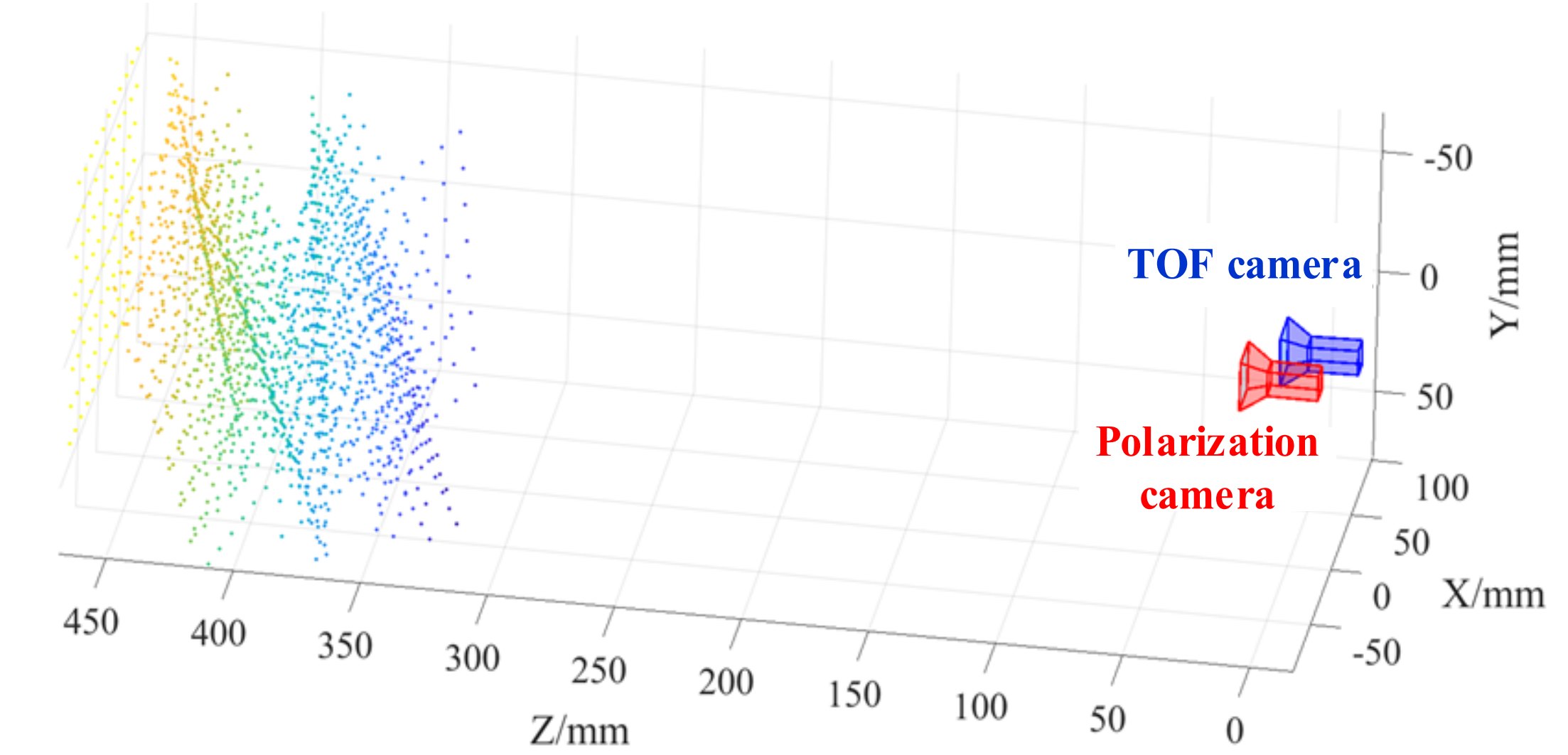

**Fig.15 Schematic diagram of the point cloud distribution of the camera and calibration plate**

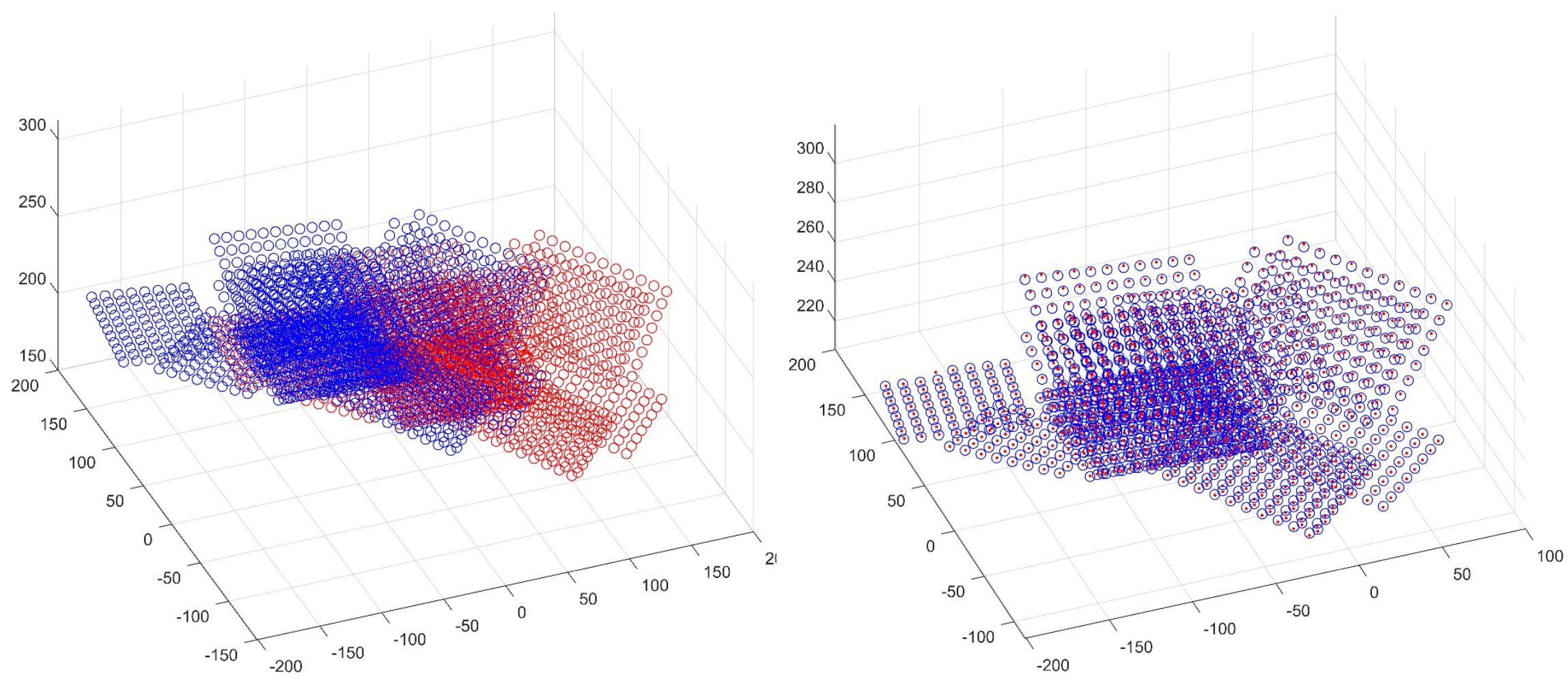

**(a)** **(b)**

**Fig.16 Schematic diagram of the registration results: (a) Before registering the calibration board corner point cloud; (b) After registering the calibration board corner point cloud; (c) Convergence curve of the calibration**

*D. Staircase positioning experiment*

1) Resolution analysis of scene reconstruction

The accuracy evaluation of 3D reconstruction based on binocular polarization and TOF is mainly based on the angle of reconstructed depth map. The reconstruction of the staircase mainly involves reconstructing the angle of the staircase steps. Therefore, for the measurement of this parameter, V-shaped plates are selected as auxiliary verification instruments to verify the accuracy of 3D reconstruction, in which the angles of V-shaped plates is 90 degrees. V-shaped plates are made of light-cured resin. The machining accuracy is 0.1mm, and it is placed 0.5m away from the camera.

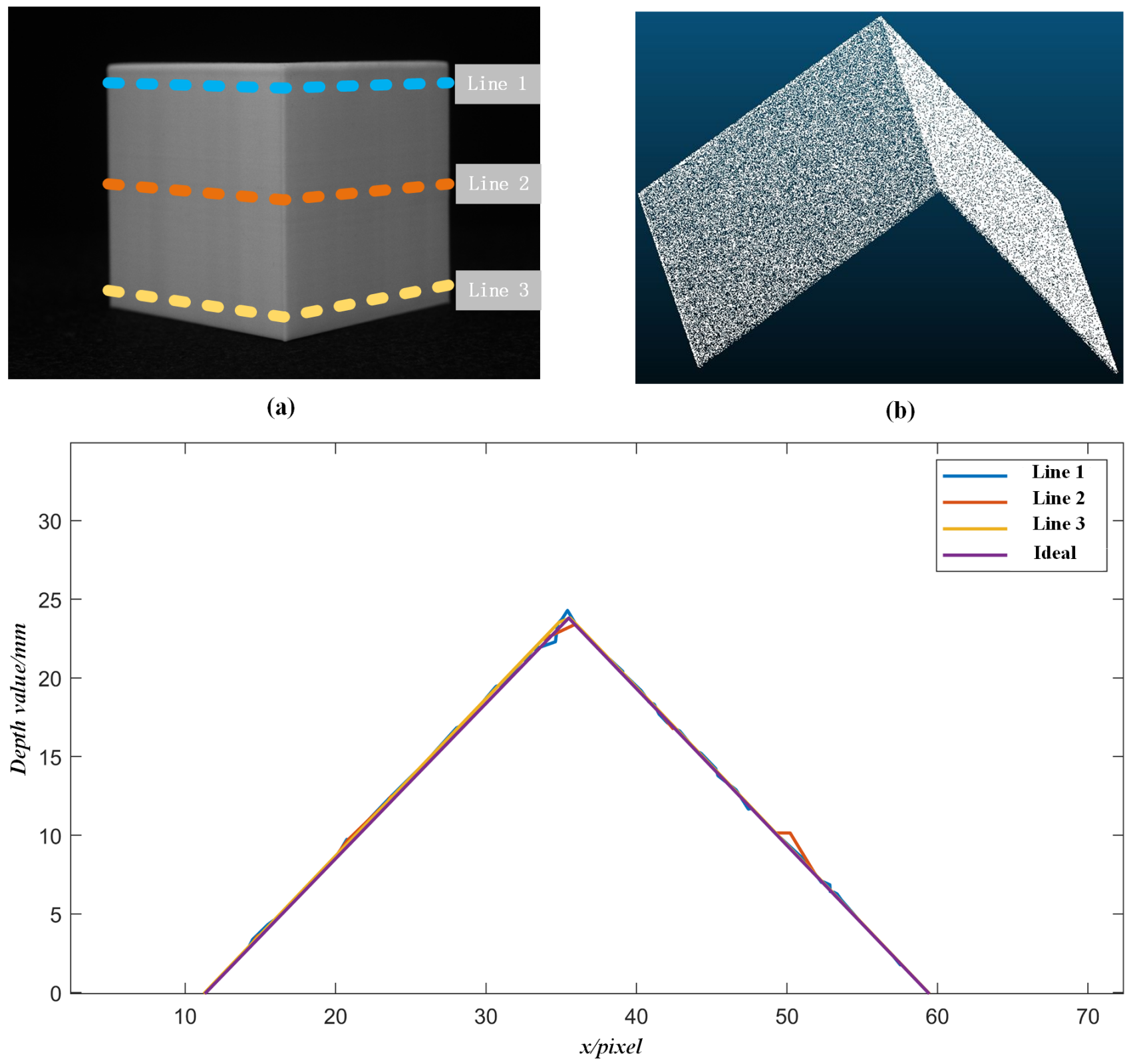

(c)

**Fig.17 90° V-plates reconstruction result**

Fig.17(b) shows the reconstruction of the V-plate in Fig.17(a). It can be seen from the figure that the reconstruction result of the V-shaped plate is relatively flat, and the reconstructed topography is close to the real object surface, which indicates that the algorithm in this paper can effectively correct the surface gradient field of the target object. Fig.17(c) shows the depth distribution at different cross sections on the V-shaped plate at different angles. From the data in the figure, it can be seen that the distribution of the reconstructed depth deviates from the ideal value in the middle of the V-shaped plate. Two factors cause this: first, the position of the slit is larger than the position of the polarization camera, which leads to a larger zenith angle and is vulnerable to large noise. In addition, the slit is the inflection point of the gradient field distribution, and the slit is the intersection of the gradient fields in two directions, which will cause distortion of the reconstruction results.

Fig.17(c) shows the distribution between the target section depth and the ideal depth of the V-shaped plate. From the image, it can be seen that the distribution trend between the target section depth and the ideal depth in Fig.17(a) is consistent. Because the polarization gradient field is calculated by the polarization optical information, it is inevitable that there will be data errors and noise in the data acquisition process. The result of the polarization gradient field can only be approximated by the gradient field, and can not be absolutely equal.

According to the ideal depth map, the linear equation $ax+by+c=0$ is fitted, and the error between the ideal depth and the actual depth is calculated by subtracting it from the actual reconstruction depth value. Based on this, the average and maximum errors of the 90° V-shaped plates are 0.46mm and 1.02mm, respectively. The standard deviation is 0.18mm, and the 95% confidence interval is 0.46±0.18 mm. The ratio of the maximum reconstruction error to the target distance is $\alpha = \frac{1.02}{500} \times 100\% \approx 0.2\%$ , greater than that of TOF camera of 1%-2%. Among them, due to the small zenith angle at the middle edge of the V-shaped plate, there is significant noise and errors at this location.

2) Comparison of Reconstruction Results

The 3D reconstruction effect of integrating binocular polarization and TOF proposed in this paper is shown in Fig.18. Experimental scene 1 and 2 are low textured indoor marble stairs and outdoor concrete stairs under different light conditions, respectively. The method proposed in this paper has been compared with polarization reconstruction [43], binocular reconstruction [44], TOF depth sensor, polarization fusion with structed light [29] and RealSense. Since we cannot get the ground truth of the staircase, we cannot compare with deep learning methods. Besides, polarization 3D reconstruction has few open-sourced algorithm, we are difficult to compare with other methods.

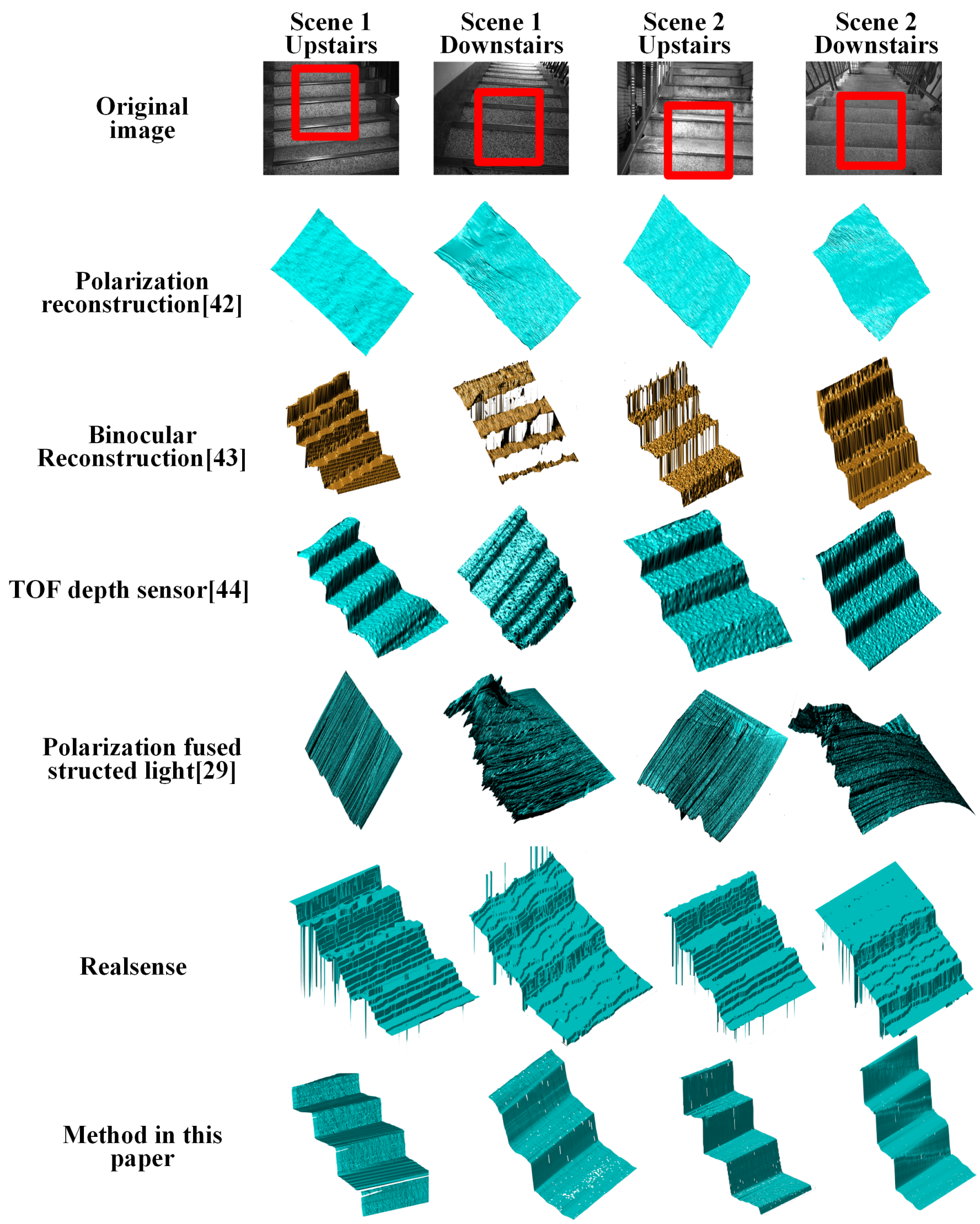

**Fig.18 Comparison of polarization 3D reconstruction results**

As shown in Fig.18, polarization 3D reconstruction results in significant distortion due to ambiguity in the direction of normal vectors, making it difficult to accurately obtain the reconstructed target morphology. Although the method of binocular 3D reconstruction can obtain a relatively correct morphology of the object surface, the details of the reconstruction results are poor and the surface is uneven due to matching errors, noise, and other effects in the stereo matching process. TOF imaging is sensitive to light due to time resolution. Under the influence of multipath effects and time resolution, the echo signal is prone to phase delay and noise during reception, resulting in significant reconstruction errors. The surface potholes in the reconstruction results are a manifestation of reconstruction errors. The polarization fusion with structed light requires static experiment environment, and the robot cannot realize this condition. The limitations of this method make it difficult to be deployed on robots. Therefore, it is difficult to achieve high-quality staircase reconstruction in outdoor environments. Due to the low texture target of the reconstructed staircase, Realsense is prone to noise when capturing complex surfaces or weak texture areas, resulting in uneven and distorted reconstruction results

As shown in the reconstruction results in the figure, since the lack of correction for gradient field ambiguity in polarization 3D reconstruction, the reconstruction results did not show the correct trend of scene morphology, presenting an overall plane. As shown in Fig.19, compared with several algorithms with relatively good reconstruction results, the algorithm in this paper portrays the details better. Specifically, its reconstruction results are closer to the object features of staircases in actual scenes, the surface is smoother, and the reconstruction distortion is smaller.

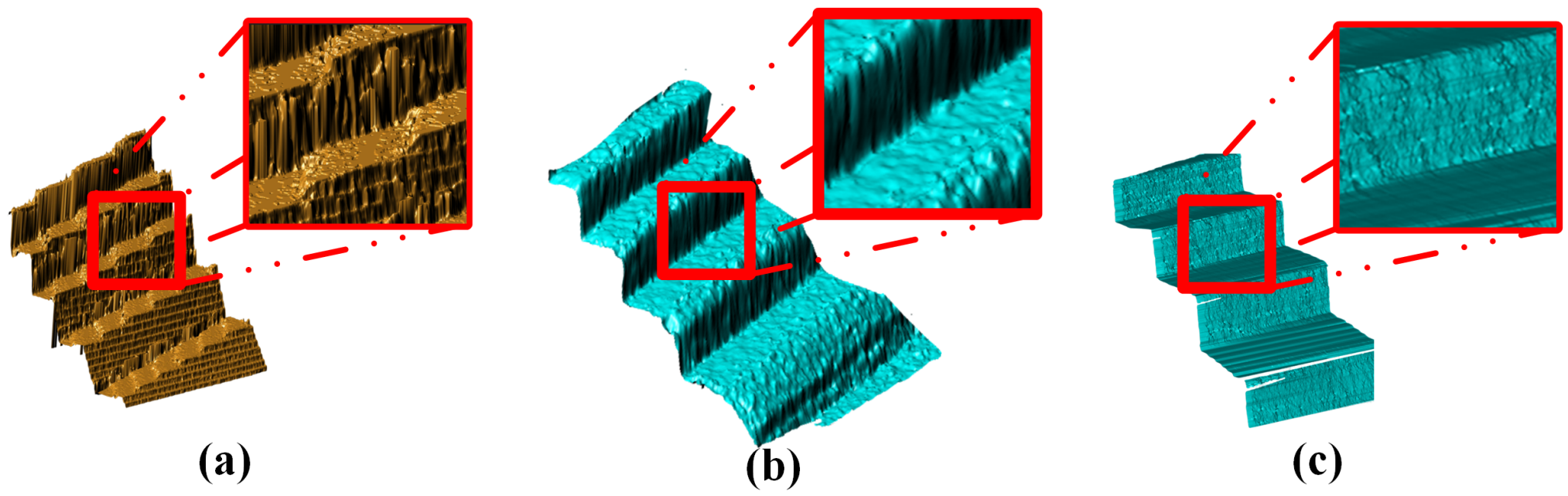

**Fig.19 The results of the reconstruction and their details are displayed: (a) Binocular reconstruction; (b) TOF depth sensor; (c) Reconstruction proposed in this paper**

## V. DISCUSSION

This paper mainly studies the perception of staircase scenes based on polarization 3D vision. A system combining scene recognition, 3D reconstruction, and point cloud processing is designed to improve recognition accuracy. Polarization vision can obtain more dimensional information than ordinary RGB or grayscale images. In reality, 3D reconstruction is based on the polarization information of the target object surface. Therefore, the fusion of binocular polarization and TOF methods can achieve the reconstruction of staircase.

This paper first focuses on the interference of reflection on target detection, and constructs a DoP image and intensity image fusion based on MLP network to improve the contrast of the input image for object detection, and proposes a point cloud segmentation based on normal vector differences for secondary recognition to improve recognition accuracy, achieving the accuracy improvement of the staircase recognition network. Secondly, the joint calibration of monocular cameras and TOF cameras was completed. In order to obtain a relatively accurate rigid transformation matrix for high-resolution monocular cameras and low resolution TOF cameras, the ICP point cloud registration method was first used to register the spatial coordinates of the checkerboard corner points. Then, the output results were optimized using an improved grey wolf optimization algorithm combining Levy flight method and chaotic mapping, ultimately achieving high-precision calibration. Finally, the stereo polarization fusion TOF method was implemented for 3D reconstruction of staircase. This 3D reconstruction method avoids the characteristics of poor mapping details, weak anti-interference ability, and being limited by time resolution and lighting of a single sensor.

Through experimental analysis and algorithm comparison, it has been verified that the scene recognition algorithm in this paper can effectively improve detection accuracy, with an accuracy of 98.7%±0.1%. Compared to using the YOLOv11 network alone, the accuracy can be improved by 3.5%. Secondly, it was verified that the heterogeneous sensor calibration accuracy of TOF monocular cameras can reach 0.5mm. Finally, based on the TOF monocular camera heterogeneous sensor calibration and comparison with classic reconstruction methods, binocular stereo matching, monocular polarization fusion depth sensors, and binocular polarization fusion photometric stereo methods, this method achieves more refined mapping and higher accuracy.

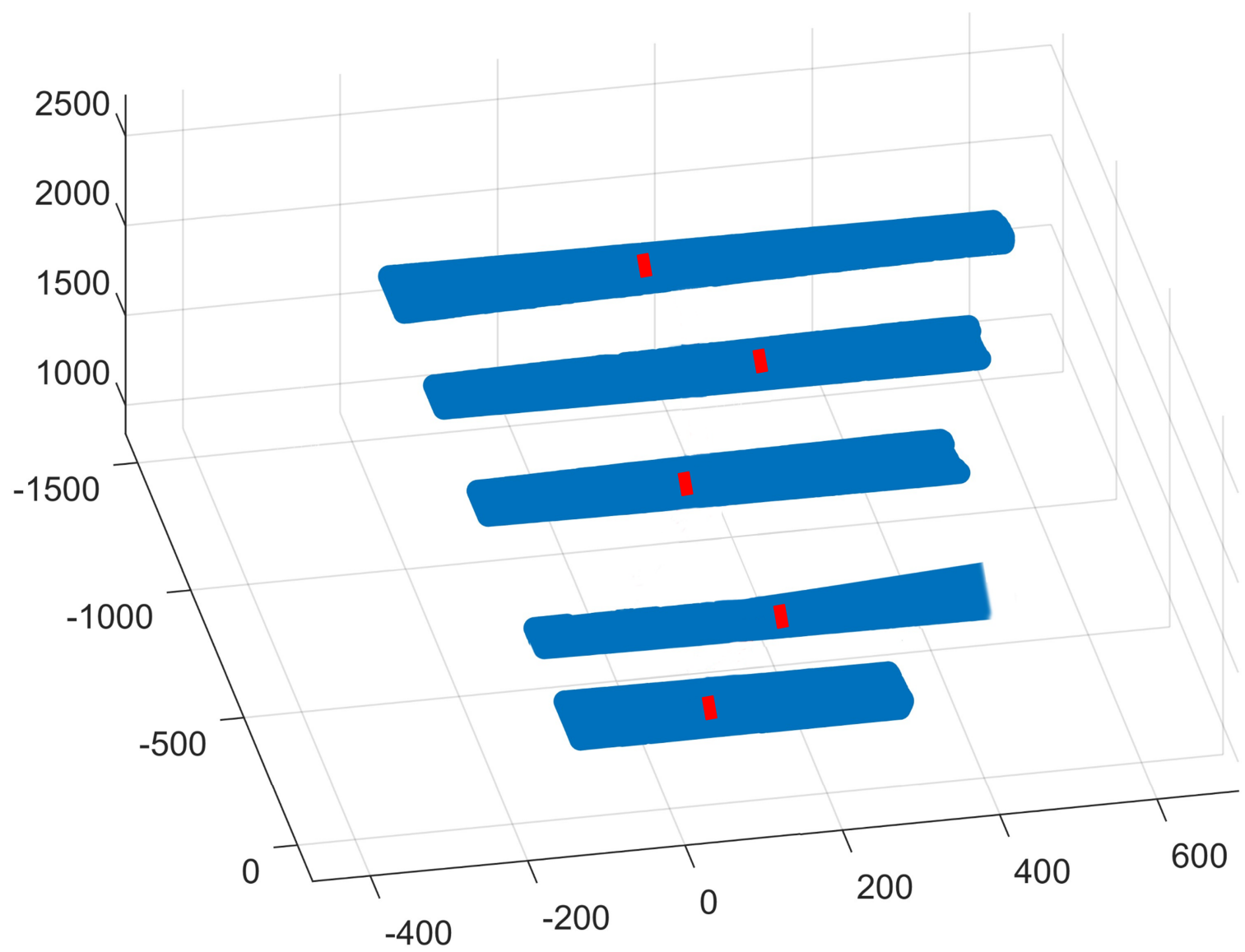

**Fig.20 Robot landing point path planning**

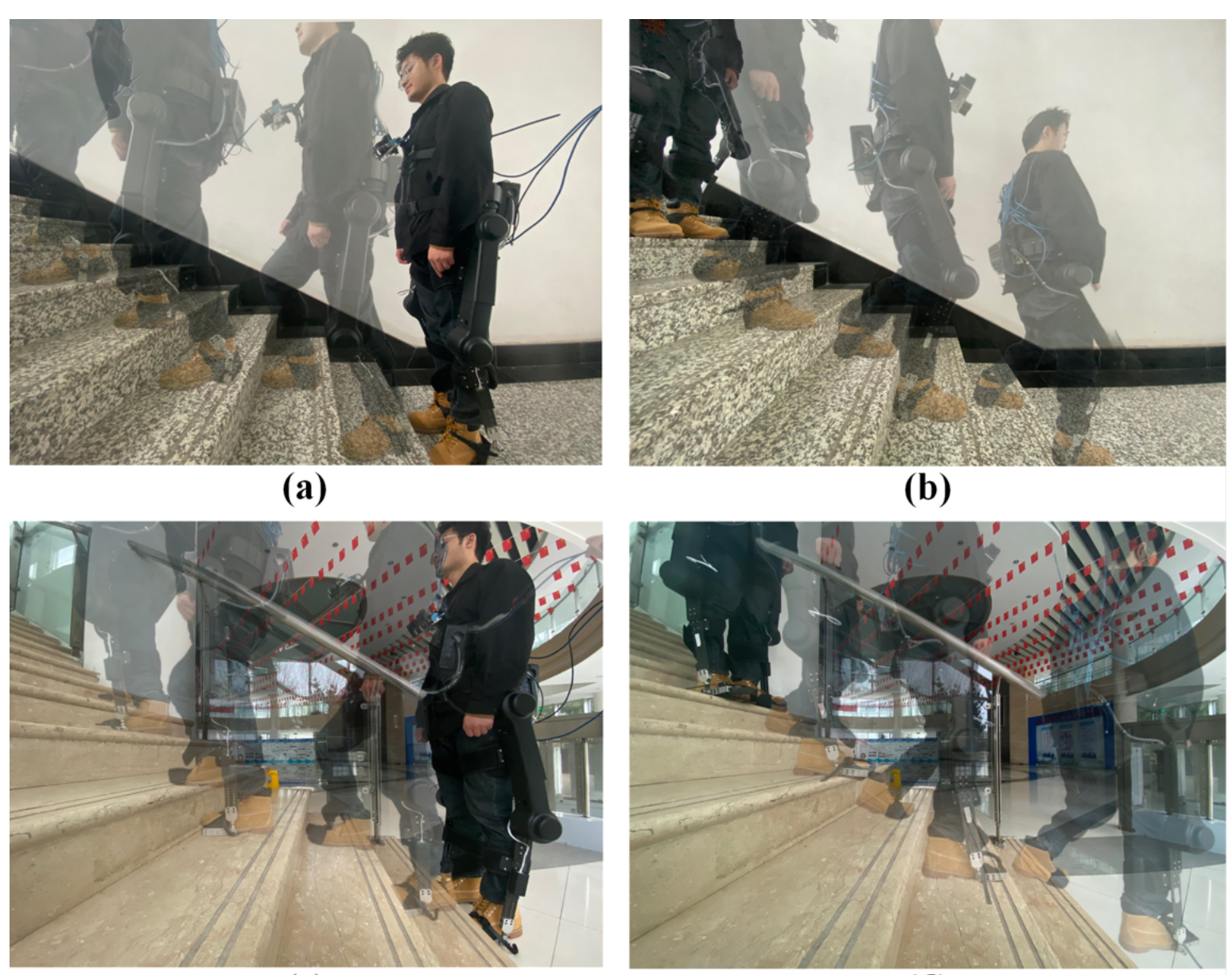

**(a)** **(b)**

**(c)** **(d)**

**Fig.21 Robot motion trajectory: (a) Scene 1: Going up the stairs; (b) Scene 1: Going down the stairs; (c) Scene 2: Going up the stairs; (d) Scene 2: Going down the stairs**

In future applications, the relevant technologies proposed in this paper can be applied to humanoid robots or exoskeleton robots for footstep planning in staircase scenes, achieving autonomous movement of humanoid robots in staircase scenes, or assisting exoskeleton robots in the field of elderly and disability assistance to help target populations traverse staircase scenes as redundant information. At present, the control of exoskeleton robots is mostly based on electromyographic signal sensors, which rely more on human motion intentions and lack the ability to perceive the scene. The introduction of visual information greatly enhances the perception ability of the scene. The fusion of photoelectric sensors and wearable sensors can further determine whether the best foot point exists in front of the current motion scene. As shown in Fig.20 and 21, the landing point planning in the staircase scene and the walking trajectory of the exoskeleton robot after introducing visual information on the staircase are shown. Due to the low robustness of the algorithm, filtering the noise that exists after segmentation leads to inconsistent step heights in the staircase scene in Fig.20. In future practical use, the segmentation algorithm still needs to be optimized.

This study focuses on the perception of staircase based on polarization-visual system including staircase recognition with polarization image enhancement and polarization 3D reconstruction, but there is still a gap with the actual application goals. To enhance the real-world applicability and deployment robustness of our system, we further discuss three key challenges and potential solutions:

(1) Sensitivity of extreme lighting conditions and mitigation strategies

Although polarization vision is less sensitive to ambient light variations compared to traditional binocular or TOF-based methods, and our system has demonstrated strong robustness under standard indoor and outdoor lighting, it still faces limitations under extreme conditions such as intense illumination, low-light environments, or sudden lighting changes. Strong lighting may saturate polarization signals, while low light increases intensity image noise, which in turn affects the MLP fusion mechanism and reduces the stability of YOLOv11-based recognition. Abrupt lighting variations can also disturb the consistency of 3D reconstruction.

To address these issues, we have initiated related studies, including polarization-based low-light image enhancement and backlight correction techniques. In future work, we aim to incorporate multi-modal imaging (e.g., infrared vision) to enhance robustness in complex lighting environments. A dynamic modality-switching mechanism based on illumination conditions will be designed to ensure key feature preservation and stable performance under extreme lighting. These efforts aim to build upon our current achievements and extend system reliability across a broader spectrum of environments.

(2) Limitations in reconstructing specular or complex surfaces

Our proposed method performs well on staircases made of marble and concrete materials. However, surfaces with strong specular reflection, such as glass and metal, remain challenging. Specular components can interfere with the polarization gradient field, resulting in inaccurate azimuth estimation, while the lack of texture in these regions can also impair stereo matching, causing missing depth information.

To overcome these challenges, we plan to introduce a reflection component separation mechanism. By analyzing differences in DoP values and polarization angle distributions, we will differentiate between diffuse and specular regions. For specular areas, TOF-based reconstruction will be prioritized to avoid polarization interference, while diffuse regions will continue to benefit from polarization–stereo fusion. In addition, we propose to apply an attention-guided refinement strategy, using TOF-based edge information to guide azimuth correction and mitigate degradation caused by texture loss. These improvements are designed to expand material generalization while preserving the current strengths of our method. In addition, multi-sensor fusion still has certain computational overhead, and in the future, lightweight network structures will be explored to reduce hardware dependencies and improve deployment efficiency on embedded devices

(3) Robustness against motion blur and frame misalignment in dynamic scene

Our current system is designed primarily for static or slow-motion environments and has already demonstrated strong performance in such cases. However, in real-world applications, rapid system motion or dynamic objects (e.g., pedestrians occluding staircases) may introduce motion blur and inter-frame misalignment, which can degrade recognition and reconstruction accuracy.

To address these dynamic challenges, we plan to integrate inertial measurement units with our vision system to achieve real-time motion compensation and scene alignment. Visual-inertial fusion, sliding-window optimization, and multi-frame feature alignment will be explored to estimate and correct for camera motion. Furthermore, optical flow-based registration and dynamic reconstruction algorithms will be investigated to improve the system's adaptability in fast-changing environments.

(4) Real time optimization of algorithms

Beyond addressing accuracy and robustness challenges, the practical deployment of our framework on mobile platforms necessitates stringent real-time performance and computational efficiency. The recognition module achieves consistent single-frame processing at 20 milliseconds (50 FPS) through optimized polarization-intensity fusion and YOLOv11 integration, ensuring real-time terrain awareness without bottlenecking the system. Crucially, the heterogeneous sensor calibration – a computationally intensive process leveraging our improved Grey Wolf Optimizer – is executed offline as a one-time operation; the resulting rigid transformation matrix is solidified in firmware and loaded during initialization, eliminating runtime calibration overhead entirely. While the core reconstruction fusion incurs higher latency, it operates strategically rather than per-frame: leveraging the relative stability of staircase geometry, high-fidelity reconstruction is triggered periodically or reactively upon significant scene changes detected by the fast recognition module, decoupling it from the navigation loop. This on-demand execution, combined with strict ROI-centric processing confining computations to detected staircase regions, effectively manages computational load. For deployment on severely resource-constrained embedded platforms (e.g., exoskeletons, micro-robots), we acknowledge the need for further acceleration of the fusion stage; future work will implement these algorithms on FPGA hardware to achieve true end-to-end real-time performance without compromising accuracy.

The above discussion highlights that our system achieves high-precision recognition and reconstruction in typical indoor and outdoor staircase scenes. While challenges remain in extending performance to extreme lighting, complex surfaces, and dynamic conditions, we have initiated corresponding research directions and identified clear development paths. Overall, the proposed method demonstrates strong innovation and practical potential in polarized vision, multi-sensor fusion, and 3D reconstruction. With these ongoing efforts, we aim to further extend the robustness and generalization of our method, while maintaining its current strengths in accuracy, efficiency, and detail preservation

## VI. Conclusion

This paper presents a comprehensive solution for robust staircase perception in robotics, addressing critical limitations of low recognition accuracy and texture dependency through a novel polarization-visual fusion framework. By establishing a paradigm of Staircase Recognition - Heterogeneous Sensor Calibration - Polarization 3D Reconstruction. In terms of staircase recognition, the accuracy of staircase scene recognition is improved by fusing DoP image and intensity image to improve the contrast of the input image and integrating 3D point cloud information. In terms of 3D reconstruction of staircase, the introduction of polarization method reduces the interference of ambient light and the dependence on object surface texture. The fusion with binocular vision and TOF reduces data holes in the reconstruction results, and corrects the problem of normal vector ambiguity in the polarization gradient field. In the fusion process of binoculars and TOF, this paper also proposes an improved grey wolf optimization algorithm for joint camera calibration. Compared with existing methods, the method proposed in this paper has higher recognition accuracy, relatively high calibration accuracy, and better texture details in the reconstructed results. Experimental validation confirms the framework's superiority in indoor/outdoor lighting/texture conditions, enabling precise foothold planning for humanoid robots and assistive devices. Future work will extend this to dynamic environments and specular surfaces through multi-modal sensing.

This work was supported by National Key R&D Program of China (2023YFB3907203), Key R&D Plan of Jiangsu Province (BE2020082-1 and BE2021016-4), Fundamental Research Funds for the Central Universities (B230201005) and Anhui Provincial Natural Science Foundation (1808085QF193).